\documentclass[10pt,twocolumn,letterpaper]{article}

\usepackage{cvpr}              

\usepackage[dvipsnames]{xcolor}

\definecolor{cvprblue}{rgb}{0.21,0.49,0.74}
\usepackage[pagebackref,breaklinks,colorlinks,citecolor=cvprblue]{hyperref}
\usepackage{multirow}

\def\paperID{3812} 
\def\confName{CVPR}
\def\confYear{2025}

\title{MultiVENT 2.0: A Massive Multilingual Benchmark \\ for Event-Centric Video Retrieval}

\author{Reno Kriz$^{\spadesuit\heartsuit}$* \hspace{2.5mm} Kate Sanders$^{\heartsuit}$* \hspace{2.5mm} David Etter$^{\spadesuit\heartsuit}$* \hspace{2.5mm} Kenton Murray$^{\spadesuit\heartsuit}$ \hspace{2.5mm} Cameron Carpenter$^{\heartsuit}$ \\
Kelly Van Ochten$^{\clubsuit}$ \hspace{2.5mm} Hannah Recknor$^{\spadesuit}$ \hspace{2.5mm} Jimena Guallar-Blasco$^{\spadesuit}$ \hspace{2.5mm} Alexander Martin $^{\heartsuit}$ \hspace{2.5mm}  \\
Sydney Johns$^{\clubsuit\diamondsuit}$ \hspace{2.5mm} Ronald Colaianni$^{\clubsuit}$ \hspace{2.5mm} Nolan King$^{\clubsuit}$ \hspace{2.5mm} Eugene Yang$^{\spadesuit\heartsuit}$ \hspace{2.5mm} Benjamin Van Durme$^{\spadesuit\heartsuit}$ \\
$^{\spadesuit}$ Human Language Technology Center of Excellence \hspace{2.5mm} $^{\heartsuit}$ Johns Hopkins University \\
$^{\clubsuit}$ SCALE Participants \hspace{2.5mm} $^{\diamondsuit}$ Virginia Tech University \\
{\tt\small \{rkriz1,ksande25,kenton\}@jhu.edu}
}

\begin{document}
\maketitle
\begin{abstract}
Efficiently retrieving and synthesizing information from large-scale multimodal collections has become a critical challenge. However, existing video retrieval datasets suffer from scope limitations, primarily focusing on matching descriptive but vague queries with small collections of professionally edited, English-centric videos. To address this gap, we introduce \textbf{MultiVENT 2.0}, a large-scale, multilingual event-centric video retrieval benchmark featuring a collection of more than 218,000 news videos and over 3,900 queries targeting specific world events. These queries specifically target information found in the visual content, audio, embedded text, and text metadata of the videos, requiring systems leverage all these sources to succeed at the task. Preliminary results show that state-of-the-art vision-language models struggle significantly with this task, and while alternative approaches show promise, they are still insufficient to adequately address this problem. These findings underscore the need for more robust multimodal retrieval systems, as effective video retrieval is a crucial step towards multimodal content understanding and generation.
\end{abstract}

\vspace{-10mm}    
\section{Introduction}
\label{sec:intro}

\begin{figure}[!t]
\centering
\includegraphics[width=0.98\linewidth]{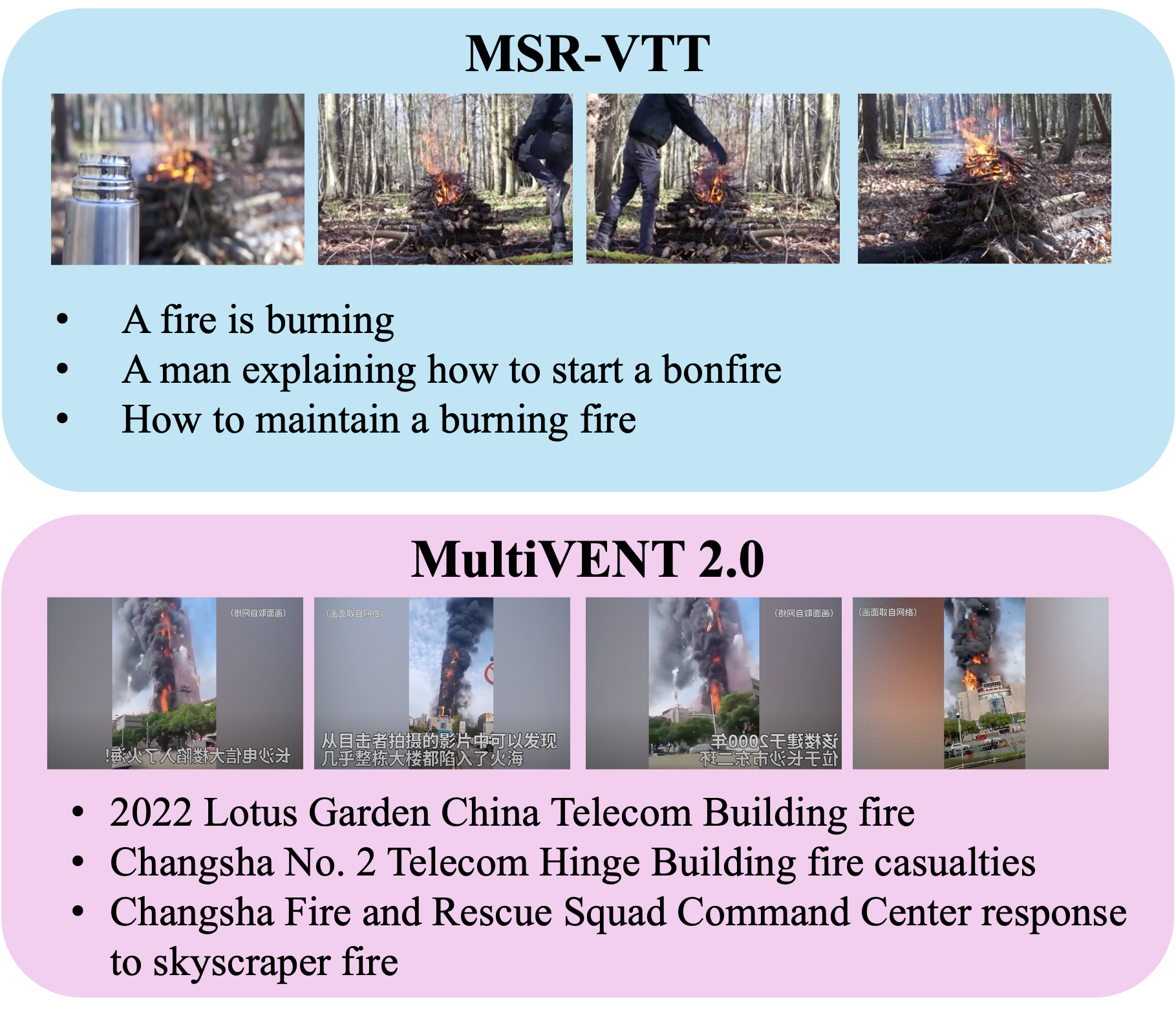}
\caption{Example query/video pairs from \textsc{MSR-VTT} \cite{xu2016msrvtt} and \textsc{MultiVENT 2.0}. \textsc{MSR-VTT} primarily contains broad descriptive queries mapped to general English-centric video clips, while \textsc{MultiVENT 2.0} targets specific current events covering 4 media formats, 6 languages, and subjects like natural disasters, politics, sports, social gatherings, and science.}
\label{fig:multivent_intro}
\end{figure}

\begin{table*}[!ht]
\small
\centering
\setlength{\tabcolsep}{4pt}
\renewcommand{\arraystretch}{0.95}
\resizebox{0.98\textwidth}{!}{%
\begin{tabular}{cccccccccc} \toprule
\textbf{Dataset} & Query Type & Query Generation & Domain & Source & \# Queries & \# Videos(/Clips) & Avg Clip Length & Multilingual \\ \midrule
ActivityNetCaptions     & Caption  & Manual & Action   & YouTube                & 100K & 20K/100K    & 120s   & no  \\ 
MSR-VTT         & Caption  & Manual & Open     & YouTube                & 200K & 7.2K/10K    & 10-30s & no  \\ 
VaTeX           & Caption  & Manual & Open/Act & Kinetics-600 (YouTube) & 825K & 42K/42K     & 10s    & no  \\
DiDeMo          & Caption  & Manual & Open     & YFCC100M (Flickr)      & 40K  & 10.5K/26.9K & 5s     & no  \\ 
LSMDC           & Caption  & Manual & Movie    & Movies                 & 128K & 202/128K    & 4-5s   & no  \\
V3C             & Metadata &   -    & Open     & Vimeo                  &  -   & 28K         & 8min   & yes \\
MSVD            & Caption  & Manual & Open     & YouTube                & 70K  & 2K/2K       & 4-10s  & no  \\
Valor-32K       & Caption  & Manual & Open     & AudioSet (Youtube)     & 32K  & 32K/32K     & 10s    & no  \\
YouCook2        & Caption  & Manual & Cooking  & YouTube                & 15.4K& 2K/15.4K    & 20s  & no  \\
TVR             & Query    & Manual & TV Shows & TV Shows               & 109K & 6/21.8K     & 76s  & no  \\
MTVR            & Query    & Manual & Movies   & TV Shows               & 218K & 6/21.8K     &  76s & yes \\
HowTo100M       & Subtitle & Manual/Automatic & Instructional & YouTube & 136M & 1.2M/136M   & 4s     & no \\
Multi-HowTo100M & Subtitle & Manual/Automatic & Instructional & YouTube & 136M & 1.2M/136M & 9s & yes \\
MultiVENT 1.0   & Caption & Manual & Open&  YouTube/Twitter & 2,400 & 2.4K & 83s & yes \\ \midrule
\textsc{MultiVENT 2.0} (ours) & Query   & Manual & Open & YouTube/Twitter & 3,900 & 218K & 145s & yes \\
\bottomrule
\end{tabular}}
\caption{Comparison of video retrieval datasets. Many of these datasets were collected as resources for general vision-language model training and did not explicitly target video retrieval. Subsequent use of these collections as a retrieval benchmark leveraged descriptive video captions or subtitles as proxy queries for English only data, which differ greatly in scope and difficulty from cross-lingual text retrieval datasets \cite{lawrie2023overviewtrec2022neuclir}. To our knowledge, \textsc{MultiVENT 2.0} is the first large-scale video dataset with queries explicitly developed for an event-focused multilingual retrieval task.}
\label{table:retrieval and pretraining datasets}
\end{table*}

While information retrieval systems for text documents have been extensively studied for decades, the landscape has shifted dramatically toward visual content, particularly videos. As of January 2024, YouTube alone likely hosts over 14 billion videos.\footnote{\href{https://www.theatlantic.com/technology/archive/2024/01/how-many-videos-youtube-research/677250/}{https://www.theatlantic.com/technology/archive/2024/01/how-many-videos-youtube-research/677250/}} Despite this explosion of visual data, there remains a dearth of research focused on the efficient retrieval, processing, and synthesis of such vast collections. Datasets that reflect the diverse array of multimodal, multilingual news sources available online could help models adapt to this shift, but existing video retrieval datasets, including MSR-VTT \cite{xu2016msrvtt}, focus primarily on semantically simple videos created for English-speaking audiences. To address this limitation, the original MultiVENT dataset was introduced, containing \textbf{Multi}lingual \textbf{V}ideos of \textbf{E}vents with aligned \textbf{N}atural \textbf{T}ext across five languages \cite{sanders2023multiventmultilingualvideosevents}. However, both \textsc{MultiVENT} (2,400 videos) and \textsc{MSR-VTT} (10,000 videos) are extremely small compared to standard information retrieval collections for text documents: In comparison, the \textsc{HC4} corpus used in the \textsc{2022 NeuCLIR TREC} shared task contains over 6 million text documents \cite{lawrie2023overviewtrec2022neuclir}. This disparity in collection size and diversity underscores the challenges in scaling video retrieval research.

To push the boundaries of video retrieval and create a more challenging and realistic task, we introduce \textsc{MultiVENT 2.0}. This dataset consists of more than 218,000 videos and over 3,900 manually-written queries targeting information about specific world events depicted within this video corpus. The videos primarily span six languages—Arabic, Chinese, English, Korean, Russian, and Spanish—and range from professionally edited news broadcasts to raw, first-person footage captured on cell phones. The events in \textsc{MultiVENT 2.0} include a wide array of types, including social and sporting events, disasters, political developments, and scientific discoveries. Queries were designed to target specific information from the visual content, audio, embedded text, and text metadata, which challenges models to effectively process and integrate information across modalities and languages and better reflects real-world retrieval scenarios.

Preliminary results show that this task presents significant challenges for current state-of-the-art vision-language models (VLMs). While specialized single-modality models show promise for queries targeting content from their respective modalities, they remain insufficient for addressing the full range of queries. These findings suggest that existing systems are not yet equipped to handle complex vision-language tasks and underscores the need for more robust multimodal systems, as effective video retrieval is a critical step toward multimodal content understanding and generation. The main contributions of this paper are:

\begin{enumerate}
\item We introduce \textsc{MultiVENT 2.0},\footnote{Data available \href{https://github.com/katesanders9/multiVENT/tree/main/data/multivent_2}{at this link}.} a large-scale multilingual video retrieval task containing more than 218,000 videos. Over 3,900 queries were crafted to target aspects of current events using visual content, audio, embedded text, and text metadata from these videos.
\item We evaluate a variety of video retrieval models, demonstrating that even state-of-the-art multimodal systems struggle in this challenging event-centric setting.
\item We baseline specialized modality-specific models; despite not being able to address the task on their own, we show their potential usefulness as components in more robust retrieval pipelines.
\end{enumerate}

\section{Related Work}
\label{sec:background}

\paragraph{Video Retrieval Datasets} There are an increasing number of video datasets, many of which support the downstream text-video retrieval task. However, there are various limitations to these existing datasets. Many of these dataset have too few videos or have videos that are too short for comprehensive evaluation of complex event understanding~\cite{sanders2024survey}. For example, the popular baseline dataset, \textsc{MSVD} \cite{chen-dolan-2011-collecting} only has 1,970 videos whose length average between 4 and 10 seconds. The larger and more recent \textsc{VaTeX} \cite{wang2020vatexlargescalehighqualitymultilingual} and \textsc{Valor-32k} datasets \cite{chen2023valorvisionaudiolanguageomniperceptionpretraining} have video clips that are on average only 10 seconds long, and the clips of the extremely large dataset of over 1 million videos, \textsc{HowTo100M} \cite{miech2019howto100mlearningtextvideoembedding}, average to only 4 seconds. Many video dataset are limited to videos of a single topic or domain, such as the \textsc{LSMDC} dataset \cite{rohrbach2016moviedescription} and \textsc{TVR} dataset \cite{lei2020tvrlargescaledatasetvideosubtitle}, which feature clips from movies and TV shows respectively, and the \textsc{YouCook2} dataset which features cooking videos~\cite{zhou2018towards}. Thus, while these datasets are quite large in terms of number of videos, they do not represent diverse events or ``real-world'' topics. Moreover, while some datasets do have non-English videos, they are limited to only one non-English language; \textsc{Multi-HowTo100M} \cite{huang2021multilingualmultimodalpretrainingzeroshot} and \textsc{MTVR} \cite{lei2021mtvrmultilingualmomentretrieval} both have only English and Chinese videos. Furthermore, despite the increase in Video datasets, very few video datasets can support event-retrieval or true multi-modal retrieval. The \textsc{DiDeMo} dataset \cite{hendricks2017localizingmomentsvideonatural} is a large dataset featuring event-centric videos. However, the creators of the \textsc{DiDeMo} dataset filtered out any non-professional edited videos, thus limiting the scope of the dataset. Additionally, captions or video descriptions are often extracted for pretraining as the ``query'' intended for the text-video retrieval task
such as \textsc{ActivityNet} captions \cite{krishna2017densecaptioningeventsvideos} and \textsc{MSR-VTT} \cite{xu2016msrvtt}. The \textsc{V3C} dataset \cite{rossetto2018v3cresearchvideo} does not use captions or subtitles but video metadata. While video descriptions or summaries are high-level representations of a video, they do no reflect human style search queries befitting of a realistic retrieval task. To more easily compare and contrast the existing datasets, we have compiled the details of said video datasets in Table \ref{table:retrieval and pretraining datasets}.   

\paragraph{Video Retrieval Methods}
Text-video retrieval is a core research area in video-language understanding \cite{cao2024rapefficienttextvideoretrieval, chen2020finegrainedvideotextretrievalhierarchical, croitoru2021teachtextcrossmodalgeneralizeddistillation, tang2024musemambaefficientmultiscale, wang2021t2vladgloballocalsequencealignment,  wang2023videotextretrievalsupervisedsparse, yang2021tacotokenawarecascadecontrastive, yu2018jointsequencefusionmodel}. There have been many proposed solutions focusing on: combining pre-extracted features from frozen text and vision encoders \cite{yu2018jointsequencefusionmodel}; transferring contrastive pre-training architectures, such as \textsc{CLIP} \cite{radford2021learningtransferablevisualmodels}, to the video domain \cite{fang2021clip2videomasteringvideotextretrieval, luo2021clip4clipempiricalstudyclip, 10.1145/3503161.3547910, xue2023clipvipadaptingpretrainedimagetext}; leveraging video specific features such as sparsity \cite{cao2024rapefficienttextvideoretrieval, lei2021moreclipbertvideoandlanguagelearning}; and aligning/fusing modalities \cite{chen2023vastvisionaudiosubtitletextomnimodalityfoundation, chen2023valorvisionaudiolanguageomniperceptionpretraining, huang2022cloverunifiedvideolanguagealignment, jin2022expectationmaximization, wang2022internvideogeneralvideofoundation,wang2024internvideo2scalingfoundationmodels,  zhu2024languagebindextendingvideolanguagepretraining}. Emphasis has been placed on aligning visual content with text, yet videos also include valuable information from text metadata, audio, and embedded text. Recently, fusion models \cite{chen2023valorvisionaudiolanguageomniperceptionpretraining, chen2023vastvisionaudiosubtitletextomnimodalityfoundation} and binding models \cite{girdhar2023imagebind, zhu2024languagebindextendingvideolanguagepretraining} have been developed to unify representations across modalities. \textsc{VAST} \cite{chen2023vastvisionaudiosubtitletextomnimodalityfoundation} notably integrates vision, audio, and subtitle information for text-video retrieval, enabling higher semantic video understanding in a unified representation. Despite these advances, many of these models are trained on datasets not designed for single-modality retrieval, let alone retrieval across multiple modalities. As a result, these systems face challenges when handling more complex retrieval tasks requiring inference across modalities and languages. This highlights the need for more comprehensive collections to facilitate the development of more robust and adaptable video retrieval systems capable of addressing the diversity of events found in real-world multimedia content.

\section{Video Collection}
\label{sec:videos}

Two significant challenges in current video retrieval research are: (1) existing collections often reward models for scene description alone, without considering the broader event context, and (2) these collections are typically limited in both size and scope. In this section, we outline the video collection process for \textsc{MultiVENT 2.0}, demonstrating how it addresses these challenges. We first review the development of MultiVENT 1.0, which initiated efforts to tackle the first issue, and then describe the expanded collection process for \textsc{MultiVENT 2.0}, a larger and more diverse dataset designed for real-world applications. The final \textsc{MultiVENT 2.0} dataset comprises over 217,000 videos, split evenly into train and test collections. 

\begin{figure*}[!ht]
\centering
\includegraphics[width=0.98\linewidth]{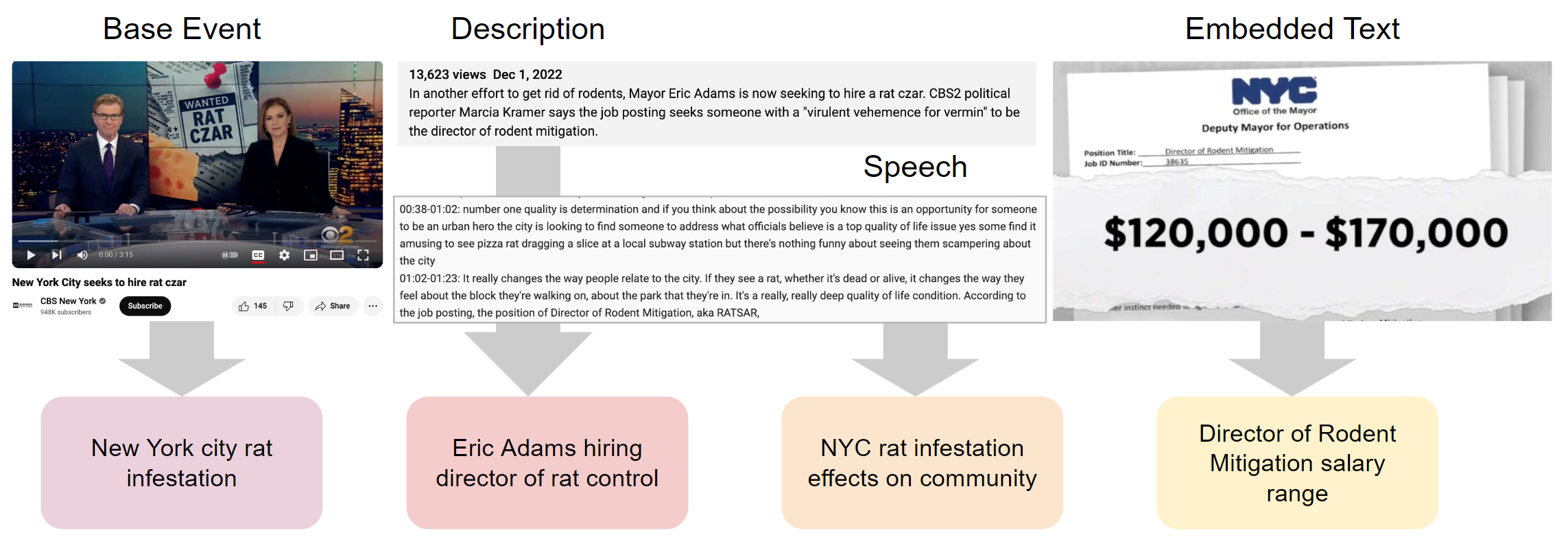}
\caption{Query creation process for event-centric videos within our distractor collection from InternVid. Annotators first create a \textbf{Base Event} query based on the primary event depicted in the video. They then write up to three additional queries focusing on specific and unique aspects of the event: the \textbf{Description} query uses only information from the human-written text description, the \textbf{Speech} query relies on spoken content from the video, and the \textbf{Embedded Text} query utilizes text visible within the video frames.}
\label{fig:multivent}
\end{figure*}

\subsection{MultiVENT 1.0 Development and Limitations}

The creation of \textsc{MultiVENT 1.0} involved several key steps: developing topics based on language- and country-specific current events, collecting relevant videos for each topic, and aligning these events with corresponding news articles. For topic selection, \citet{sanders2023multiventmultilingualvideosevents} utilized Google Trends statistics from countries with the largest populations of speakers for each target language to identify visually salient current events. After filtering to ensure adequate online video coverage, this process produced 2,396 videos spanning 255 events.

This principled approach resulted in an important dataset, filling a unique role as a targeted, event-centric video retrieval collection. However, while \textsc{MultiVENT 1.0} was a significant step forward, its small size, especially the limited distractor set, limits its applicability for large-scale multimodal retrieval research. As shown in Table \ref{table:retrieval and pretraining datasets}, this issue is common across many video retrieval tasks, and stands in stark contrast to text retrieval benchmarks, which typically consist of much larger collections. For example, the \textsc{HC4} dataset used in the \textsc{2022 NeuCLIR TREC} shared task contains over 6 million text documents \cite{lawrie2023overviewtrec2022neuclir}. This disparity in collection size underscores the need for a larger and more diverse video retrieval corpus. 

\subsection{Expanded MultiVENT 2.0 Video Collection Process}

To create an expanded collection for a more realistic, large-scale video retrieval task, we augment the original \textsc{MultiVENT} dataset with videos from \textsc{InternVid}, a corpus containing more than seven million YouTube videos and over 760,000 hours of content \cite{wang2024internvidlargescalevideotextdataset}. InternVid covers a superset of target languages and event categories than those found in \textsc{MultiVENT 1.0}. While not all \textsc{InternVid} categories are event-based, the corpus still includes a significant amount of event-centric content, particularly in the political and disaster domains.

For \textsc{MultiVENT 2.0}, we extract a large subset of videos from \textsc{InternVid}, filtering out those longer than five minutes. This process yields approximately 40,000 videos for each of the \textsc{MultiVENT 1.0} target languages: Arabic, Chinese, English, Korean, and Russian. To introduce greater variability at test time, we also include a smaller set of videos from the Spanish and ``Unknown" language categories, ensuring that systems handle both new primary languages and a long tail of low-resource languages.

After merging the expanded collection with \textsc{MultiVENT 1.0}, we compiled a final dataset of over 217,000 videos, consisting of 108,500 videos for training (\textsc{MultiVENT Train}) and 109,800 for testing (\textsc{MultiVENT Test}). All videos from \textsc{MultiVENT 1.0} are solely found in the evaluation set, and any duplicates between the training and test collections were removed. Given the scale of the dataset, rapidly comparing models and system variations can be challenging. To mitigate this and aid model tuning, we offer a subset of 2,000 videos from the training set, referred to as \textsc{MultiVENT Train-2k}.

\section{Query Creation}
\label{sec:query}

As mentioned in Section~\ref{sec:videos}, prior collections have focused more on matching descriptive aspects within videos. Accordingly, search queries have either been relatively short and vague (e.g., \textit{a black and white horse runs around}) or derived directly from a video’s metadata, such as a YouTube description. \textsc{MultiVENT 1.0} largely followed this precedent, relying on the available metadata for each video. However, these approaches create a disconnect with modern text retrieval practices, where concise, search-engine-style English queries are increasingly used to match multilingual documents \cite{lawrie2023overviewtrec2022neuclir}. To bridge this gap, we develop a novel two-pronged approach for creating event-centric video queries: one that leverages additional fine-grained event annotations available with \textsc{MultiVENT 1.0} videos \cite{sanders2023multiventmultilingualvideosevents}, and another that relies solely on the videos themselves.

For this task, we recruit professional linguists with expertise across the six primary languages targeted in \textsc{MultiVENT 2.0} to develop the queries. Each annotator underwent training via a tutorial task with detailed instructions. Afterward, we provide one-on-one feedback on their performance and closely monitor their initial annotations, offering additional guidance as needed. Finally, an annotation lead conducts a quality control check to ensure the conciseness and specificity of each query/video pair.

\begin{figure*}[htb]
    \centering
    \begin{subfigure}[b]{0.46\textwidth}
        \centering
        \includegraphics[width=\textwidth]{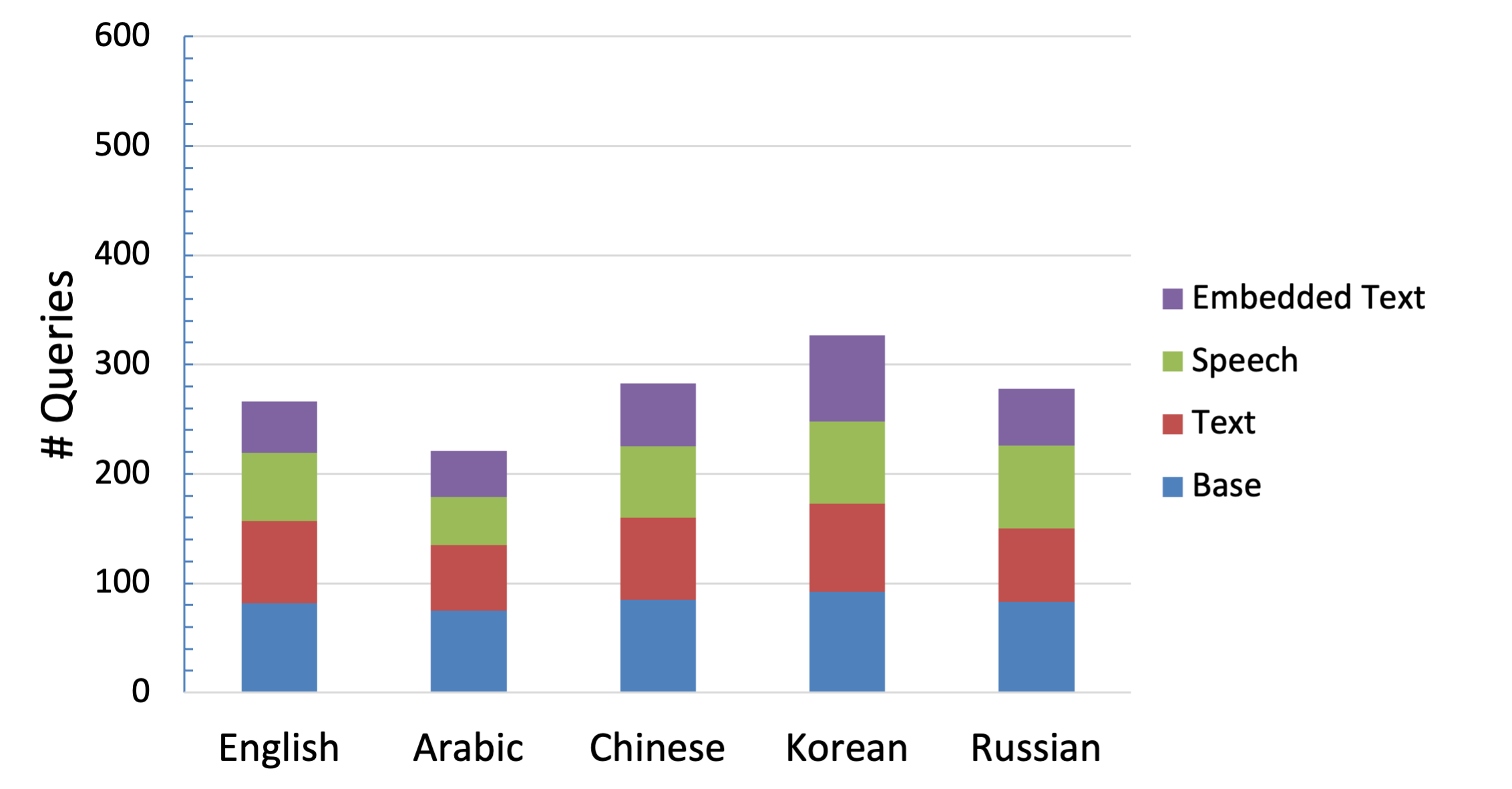}
        \caption{}
        \label{fig:language_train}
    \end{subfigure}
    \hfill
    \begin{subfigure}[b]{0.46\textwidth}
        \centering
        \includegraphics[width=\textwidth]{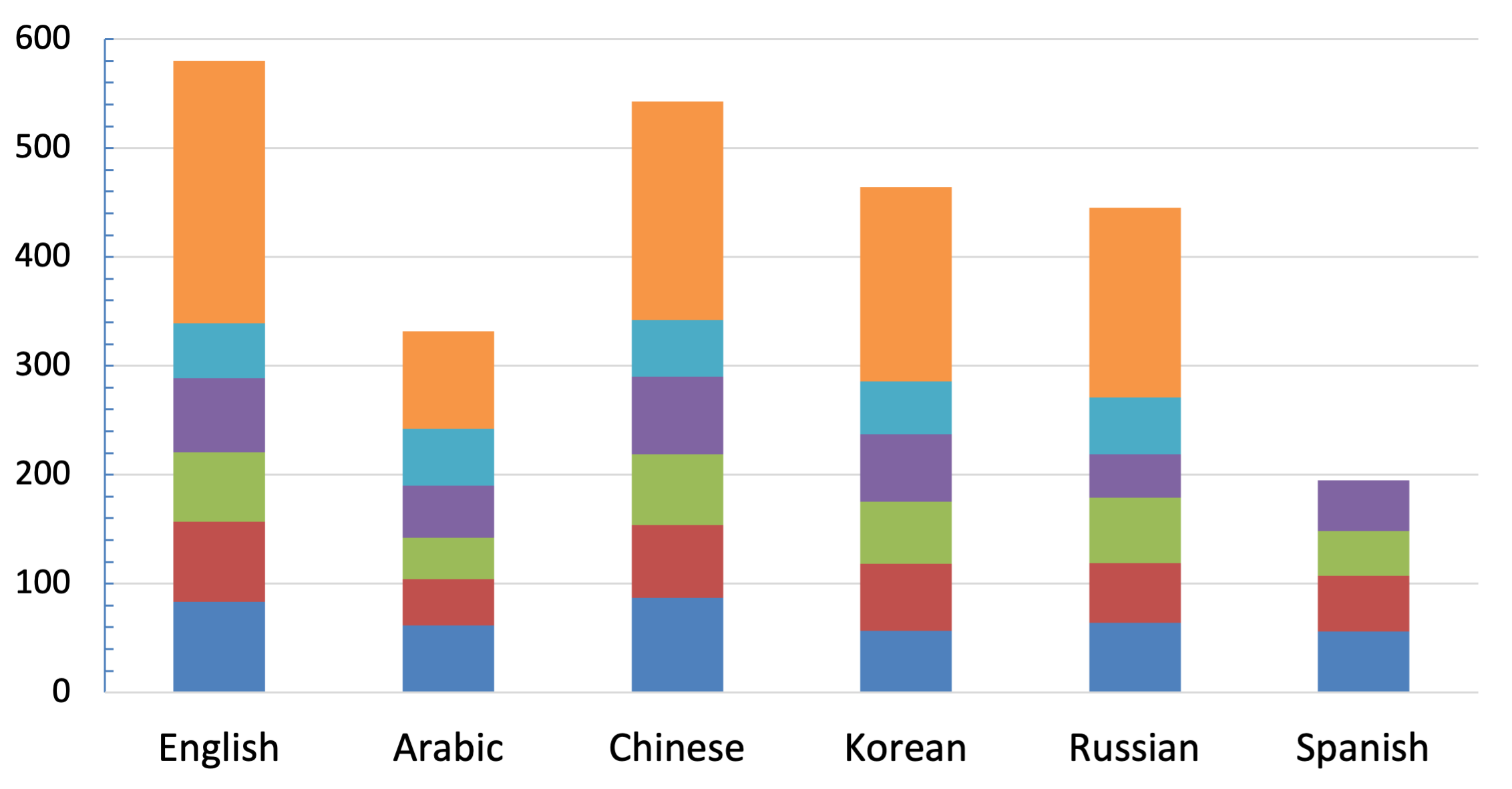}
        \caption{}
        \label{fig:language_test}
    \end{subfigure}

    \vskip\baselineskip 

    \begin{subfigure}[b]{0.46\textwidth}
        \centering
        \includegraphics[width=\textwidth]{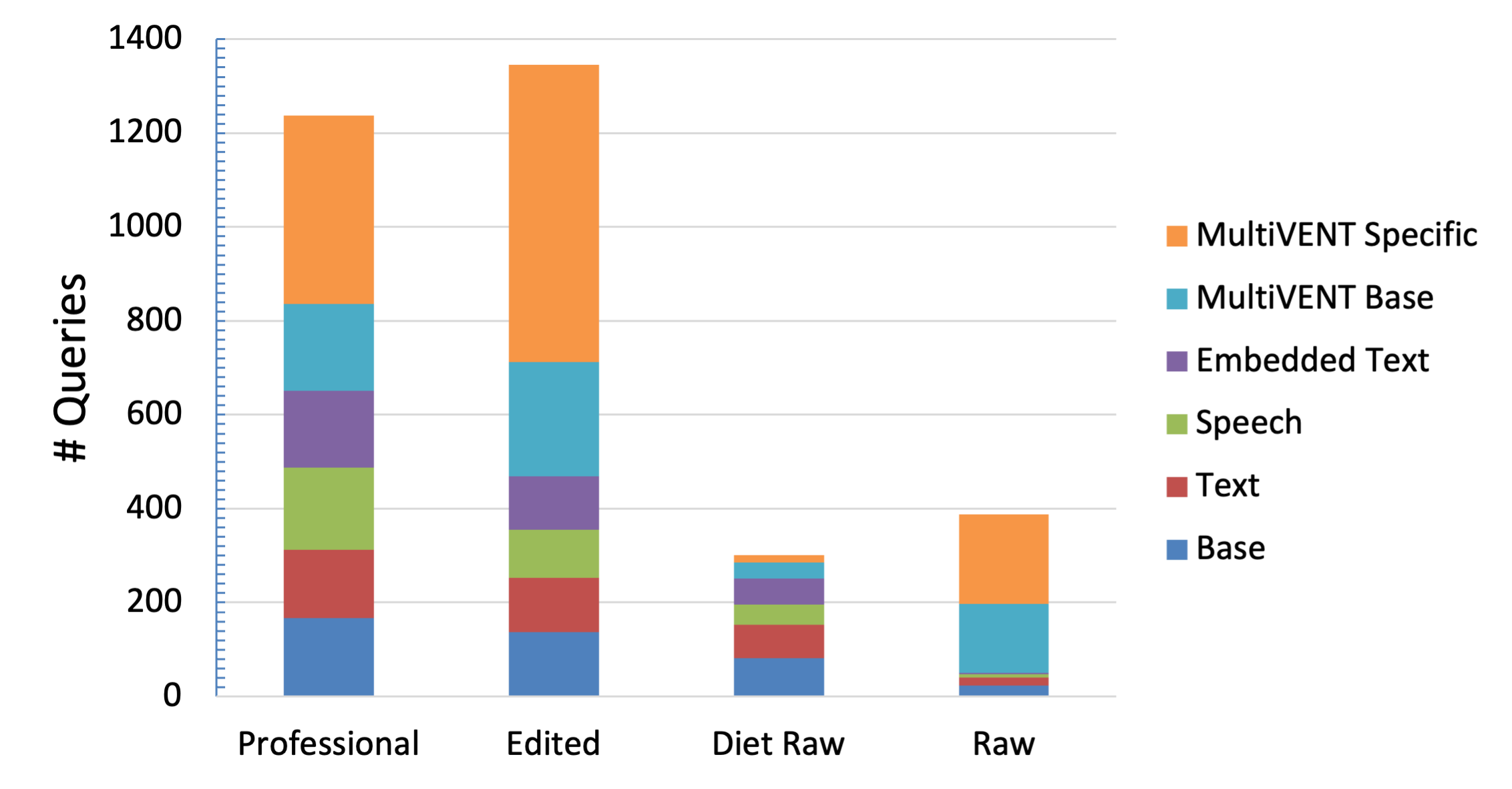}
        \caption{}
        \label{fig:video_type}
    \end{subfigure}
    \hfill
    \begin{subfigure}[b]{0.46\textwidth}
        \centering
        \includegraphics[width=\textwidth]{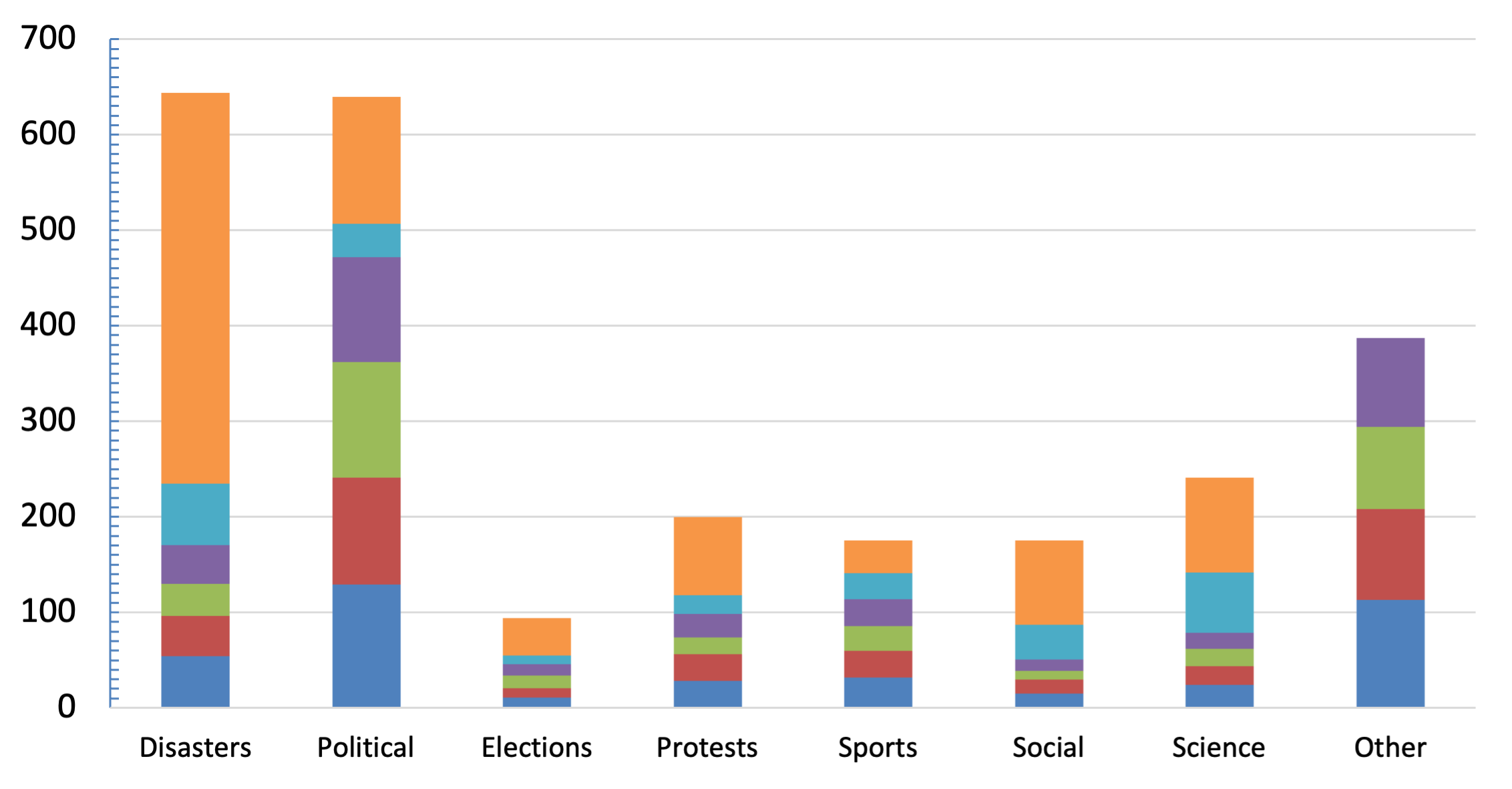}
        \caption{}
        \label{fig:event_type}
    \end{subfigure}
    
    \caption{Breakdowns of of the number of queries mapped to relevant videos. Figure~\ref{fig:language_train} shows that \textsc{MultiVENT Train} contains queries targeting the five primary languages from \textsc{MultiVENT 1.0}. On the other hand, in Figure~\ref{fig:language_test} we see that \textsc{MultiVENT Test} adds queries targeting Spanish events to challenge systems' multilingual robustness. Figure~\ref{fig:video_type} shows that \textsc{MultiVENT 2.0} targets videos ranging from professional news broadcasts to raw first-person footage of events. Finally, as seen in Figure~\ref{fig:event_type}, events in \textsc{MultiVENT 2.0} generally map to the same categories as MultiVENT 1.0, with a long tail of infrequent event types.}
    \label{fig:multivent_breakdowns}
\end{figure*}

\subsection{Updated Query Creation for MultiVENT 1.0} \label{sec:multivent1_query}

In \textsc{MultiVENT 1.0}, the text descriptions accompanying each video were initially repurposed as queries. However, this dataset also provides a foundation for crafting more targeted, event-centric queries, as each of the 255 current events in the collection is aligned with an article in its original language, along with an English version if the source language is not English. Many of these English articles come from Wikipedia, where titles are typically concise and specific descriptions of the broader event. For instance, the article aligned with the event shown in Figure~\ref{fig:multivent_intro} is titled \textit{2022 Lotus Garden China Telecom Building fire}. Annotators follow this style to manually create a base query for each current event in \textsc{MultiVENT 1.0}, henceforth referred to as \textsc{MultiVENT Base} event queries.

While retrieving videos about overarching events is a valuable step forward, research in text information extraction (IE) has shown that individual events often encompass multiple distinct aspects, which are unlikely to be fully captured in a single piece of media. Thus, to further challenge retrieval models, annotators develop queries focused on specific, unique aspects of each event. This is facilitated by leveraging annotations from \textsc{MultiVENT-Grounded}~\cite{sanders2024grounding}, a collection of 1,200 \textsc{MultiVENT 1.0} videos containing fine-grained event annotations. These annotations, guided by adapted \textsc{FrameNet} event templates, include text description spans, video time intervals, and spatial bounding boxes at the frame level, and address IE-based questions such as \textit{Where did the disaster occur?}, \textit{Who was affected by the disaster?}, and \textit{Who responded to the disaster?} Based on these annotations, annotators create 884 \textsc{MultiVENT Specific} queries, each targeting videos that highlight distinct aspects of the corresponding event.

\subsection{InternVid Query Creation} \label{sec:internvid_query}

The previous section focuses on the retrieval of \textsc{MultiVENT 1.0} videos, despite the majority of our collection coming from InternVid. This raises the risk that part of the task could devolve into a binary classification problem, where simply distinguishing between MultiVENT and InternVid videos significantly reduces the task's complexity. To mitigate this, we tasked annotators with writing queries targeting new events found within the larger InternVid collection. Since there was no guarantee that InternVid videos contained event-based content, we focused annotation efforts on videos from the News \& Politics and Sports categories. Annotators were first asked to confirm whether a video contained event-based content. If confirmed, they write a \textbf{Base Event} query in the same style as those described in Section~\ref{sec:multivent1_query}. Additionally, annotators searched the web for an article related to the event.

Next, annotators develop up to three additional queries, each focusing on specific and unique aspects of the same event. For these queries, annotators were asked to rely on partial information about a video from a single modality: the \textbf{Description} query uses only the human-written YouTube description of the video; the \textbf{Speech} query uses any spoken content from the video, with Whisper-generated speech transcripts provided for assistance~\cite{radford2022whisper}; and the \textbf{Embedded Text} query focused solely on text directly visible in the video frames, with output provided from a multilingual OCR system~\cite{etter2023hybrid}. Note that not all videos contain useful or unique information across all modalities; in such cases, no query is created. This process was applied to both the train and test collections, resulting in 1,417 additional test queries and 1,375 train queries.

\subsection{Query and Video Breakdowns} \label{sec:breakdowns}

To gain insight into the challenges posed by different types of video/query pairs, annotators are asked several additional questions about the videos during the query creation process. First, we ask them to confirm the video's primary language. While \textsc{InternVid} ostensibly provides this information, the labels are sometimes incorrect; this is particularly common in English-labeled videos, where approximately 20\% of the videos were in other languages. Figure~\ref{fig:language_train} shows the distribution of queries targeting videos from each of the five original \textsc{MultiVENT 1.0} languages within \textsc{MultiVENT Train}, while Figure~\ref{fig:language_test} illustrates the same for \textsc{MultiVENT Test}. Notably, we add Spanish as an additional primary language in the test collection to evaluate the multilingual robustness of video retrieval systems.

Additionally, linguists group events into one of the seven event types defined in \textsc{MultiVENT 1.0}: natural disasters (\textbf{Disasters}), political elections (\textbf{Elections}), \textbf{Protests}, other political developments (\textbf{Political}), sporting events (\textbf{Sports}), social events (\textbf{Social Events}), and scientific or technological discoveries (\textbf{Science})~\cite{sanders2023multiventmultilingualvideosevents}. An \textbf{Other} category is also provided for events that did not fit into these types. Figure~\ref{fig:event_type} depicts the event type breakdown. Two notable aspects include the relative abundance of MultiVENT-specific queries for disaster events, largely due to \textsc{MultiVENT-Grounded} annotations prioritizing this category~\cite{sanders2024grounding}, and the predominance of Political events, which reflects the news and politics videos comprising the majority of event-based content within the InternVid collection. The Other category includes diseases, police incidents, and man-made disasters, among other infrequent event types.

Finally, we ask annotators to categorize videos into three general types: \textbf{Professional} news broadcasts, i.e., videos with reporters and/or traditional news chyron; non-professional \textbf{Edited} videos, i.e., videos featuring multiple spliced clips, visual effects, or superimposed graphics; and \textbf{Raw} footage, i.e., single-stream videos of events as they happen, typically captured on a mobile device. Given the widespread availability of video editing software, it is very easy for more people to make light edits prior to uploading. To differentiate between this and "true" raw content, we add a fourth category, \textbf{Diet Raw}, for single-stream videos with minimal text and speech overlays. Figure 3c shows the distribution of queries for relevant videos across these types. From this, we can see that true raw content is relatively scarce in InternVid, reinforcing the importance of the original MultiVENT collection. Note that many queries targeting MultiVENT 1.0 events are mapped to multiple relevant videos across different video types.

\section{Relevance Judgment Annotation}
\label{sec:relevance}

After the initial query creation process, we have a total of 6,068 annotated video/query pairs for \textsc{MultiVENT-Test}. In many cases, queries are only mapped to a single relevant video, and all other non-judged videos are assumed to be not relevant. This presents a challenge because, despite the targeted nature of our queries, it is likely that the distractor set contains additional relevant videos, given the size and diversity of the test collection. Furthermore, some videos may be partially relevant to a query, or relevant to some parts but not all, while others' relevance may be unclear due to a lack of context or inherent video ambiguity.

\setlength{\belowcaptionskip}{-1em}
\begin{table}[ht!]
\small
\centering
\setlength{\tabcolsep}{4pt}
\renewcommand{\arraystretch}{0.95}
\resizebox{0.45\textwidth}{!}{%
\begin{tabular}{c|ccc} \toprule
Rank & Very Relevant & Somewhat Relevant & Not Relevant \\ \midrule
1 & 43\% & 13\% & 44\% \\
2 & 28\% & 15\% & 57\% \\
3 & 28\% & 19\% & 53\% \\
4 & 20\% & 21\% & 59\% \\
5 & 18\% & 18\% & 65\% \\
6 & 14\% & 18\% & 69\% \\
7 & 14\% & 17\% & 70\% \\
8 & 10\% & 11\% & 78\% \\
9 & 10\% & 15\% & 75\% \\
10 & 9\% & 13\% & 79\% \\ \bottomrule
\end{tabular}}
\caption{Percentage of initially un-judged video/query pairs re-annotated as not relevant, somewhat relevant, and very relevant. The \textit{Rank} indicates the ranking of a video for the corresponding query, as judged by our best baseline model.}
\label{table:relevance}
\end{table}

To address this, we re-train the same pool of professional linguists from the query creation task in Section 4 to now judge the relevance of previously unseen query/video pairs. For each candidate video, annotators are asked to classify it as \textit{not relevant}, \textit{possibly relevant}, \textit{partially relevant}, or \textit{very relevant} to a query. Given the potentially vast scope of this annotation task, we limit the videos to those ranked in the top 10 by multilingual CLIP (\textsc{mCLIP}), a pre-trained vision-language model~\cite{carlsson2022cross} and the strongest-performing model on \textsc{MultiVENT 1.0}~\cite{sanders2023multiventmultilingualvideosevents}. After removing video/query pairs already judged, we prioritize judging the highest-ranked videos, and streamline the process by grouping query/video pairs where the same video was ranked in the top 10 for multiple queries. Through this process, we collect an additional 4,396 gold relevance judgments. For evaluation purposes, we condense the middle two categories into a single category labeled \textit{somewhat relevant}.

\setlength{\belowcaptionskip}{0.1em}
\begin{table*}[ht!]
\small
\centering
\setlength{\tabcolsep}{4pt}
\renewcommand{\arraystretch}{0.95}
\resizebox{0.9\textwidth}{!}{%
\begin{tabular}{cc|c|ccccc} \toprule
\multirow{2}{*}{\textbf{Modality}} & \multirow{2}{*}{\textbf{Model}} & \textbf{MSR-VTT} & \multicolumn{5}{c}{\textbf{MultiVENT 2.0}} \\
 & & R@10 & R@10 & R@100 & MRR & mAP & nDCG@10 \\ \midrule
Vision & mCLIP & 0.827 & 0.333 & 0.603 & 0.429 & 0.261 & 0.303 \\ 
OCR & ICDAR $\rightarrow$ mCLIP & - & 0.227 & 0.374 & 0.363 & 0.166 & 0.217 \\ 
Speech & Whisper $\rightarrow$ mCLIP & - & 0.290 & 0.450 & 0.417 & 0.212 & 0.267 \\
Text & Description $\rightarrow$ mCLIP & - & 0.293 & 0.491 & 0.445 & 0.228 & 0.284 \\ \midrule
Vision & InternVideo2.0 & \textbf{0.851} & 0.004 & 0.018 & 0.018 & 0.003 & 0.005 \\
All & VAST & 0.739 & 0.118 & 0.118 & 0.198 & 0.080 & 0.116 \\
All & LanguageBind & 0.787 & \textbf{0.355} & \textbf{0.620} & \textbf{0.443} & \textbf{0.283} & \textbf{0.324} \\
\bottomrule
\end{tabular}}
\caption{Performance of pre-trained multimodal benchmark models and single-modality pipeline systems. We can see that prior state-of-the-art video retrieval systems struggle significantly with the increase in query complexity and collection size/diversity of \textsc{MultiVENT 2.0}. \textsc{VAST} is run with vision input only on \textsc{MultiVENT 2.0}, as that achieves highest performance on this task.}
\label{tab:baselines}
\end{table*}

Given these graded relevance categories, it follows that any video judged as very relevant for a \textbf{MultiVENT Base} query must be at least somewhat relevant for all associated \textbf{MultiVENT Specific} queries. To ensure systems receive appropriate credit, we apply this logic to add an additional 5,653 silver relevance judgments. This results in a final updated set of 16,116 judged video/query pairs. Prior to this annotation effort, on average only 16\% of the videos ranked in the top 10 by multilingual CLIP had an associated relevance judgment (\textit{Judged@10}); with the added judgments, the \textit{Judged@10} has now increased to 39\%. For evaluation metrics that account for graded relevance, such as normalized Discounted Cumulative Gain (nDCG), we apply a 0-1-3 scale: \textit{Very relevant} video/query pairs are scored as 3, \textit{Somewhat Relevant} pairs as 1, and \textit{Not Relevant} or non-judged pairs as 0.

\section{Baselines and Results}
\label{sec:baselines}
Jointly pre-trained vision-language models (VLMs) have recently achieved state-of-the-art results on various downstream video understanding tasks. These models typically consist of separate single-modality encoders, with a mechanism for fusing multimodal embeddings during training. In this section, we consider several prominent VLMs as baselines for video retrieval. Each of the following VLMs utilize a version of CLIP for the vision component, with the notable exception of \textsc{InternVid} using a Vision Transformer (ViT) \cite{dosovitskiy2021imageworth16x16words}.

\begin{table*}[!ht]
    \centering
    \begin{subtable}{0.98\textwidth} 
        \small
        \centering
        \begin{tabular}{cc|cccccc} \toprule
            \multirow{2}{*}{\textbf{Modality}} & \multirow{2}{*}{\textbf{Model}} & \multicolumn{6}{c}{\textbf{Performance by Language}} \\
            & & Arabic & Chinese & English & Korean & Russian & Spanish \\ \midrule
            Vision & mCLIP & \textbf{0.323} & 0.074 & \textbf{0.331} & 0.070 & \textbf{0.247} & 0.272 \\ 
            OCR & ICDAR $\rightarrow$ mCLIP & 0.320 & 0.083 & 0.171 & 0.107 & 0.104 & 0.289 \\ 
            Speech & Whisper $\rightarrow$ mCLIP & 0.232 & \textbf{0.130} & 0.203 & 0.114 & 0.224 & \textbf{0.320} \\ 
            Text & Description $\rightarrow$ mCLIP & 0.246 & 0.127 & 0.199 & \textbf{0.137} & 0.242 & 0.281 \\ \bottomrule
        \end{tabular}
        \caption{}
        \label{tab:language}
    \end{subtable}
    \vspace{0.2cm}
    \begin{subtable}{0.98\textwidth} 
        \small
        \centering
        \begin{tabular}{cc|cccc} \toprule
            \multirow{2}{*}{\textbf{Modality}} & \multirow{2}{*}{\textbf{Model}} & \multicolumn{4}{c}{\textbf{Performance by Video Type}} \\
            & & Professional & Edited & Diet Raw & True Raw \\ \midrule
            Vision & mCLIP & 0.201 & \textbf{0.276} & \textbf{0.152} & 0.198\\ 
            OCR & ICDAR $\rightarrow$ mCLIP & 0.184 & 0.176 & 0.094 & 0.006 \\ 
            Speech & Whisper $\rightarrow$ mCLIP & \textbf{0.318} & 0.154 & 0.011 & 0.021 \\ 
            Text & Description $\rightarrow$ mCLIP & 0.206 & 0.245 & 0.142 & \textbf{0.307} \\ \bottomrule 
        \end{tabular}
        \caption{}
        \label{tab:video_type}
    \end{subtable}
    \vspace{0.2cm} 
    \begin{subtable}{0.98\textwidth} 
        \small
        \centering
        \begin{tabular}{cc|cccccc} \toprule
            \textbf{Modality} & \textbf{Model} & \multicolumn{6}{c}{\textbf{Performance by Query Type}} \\
            & & MultiVENT-Base & Multivent-Specific & Base & Text & Speech & Embedded Text \\ \midrule
            Vision & mCLIP & \textbf{0.411} & 0.331 & \textbf{0.294} & \textbf{0.294} & 0.227 & 0.232 \\
            OCR & ICDAR $\rightarrow$ mCLIP & 0.254 & 0.198 & 0.219 & 0.232 & 0.162 & \textbf{0.245} \\ 
            Speech & Whisper $\rightarrow$ mCLIP & 0.268 & 0.225 & 0.282 & 0.256 & \textbf{0.306} & 0.230 \\ 
            Text & Description $\rightarrow$ mCLIP & 0.403 & \textbf{0.359} & 0.219 & 0.282 & 0.160 & 0.153 \\  \midrule 
        \end{tabular}
        \caption{}
        \label{tab:query_type}
    \end{subtable}
    \vspace{0.2cm} 
    \begin{subtable}{0.98\textwidth} 
        \small
        \centering
        \begin{tabular}{cc|cccccccc} \toprule
            \multirow{2}{*}{\textbf{Modality}} & \multirow{2}{*}{\textbf{Model}} & \multicolumn{8}{c}{\textbf{Performance by Event Type}} \\
            & & Disaster & Election & Protest & Political & Sports & Social & Science & Other \\ \midrule
            Vision & mCLIP & 0.334 & \textbf{0.215} & \textbf{0.310} & \textbf{0.334} & \textbf{0.441} & \textbf{0.486} & \textbf{0.373} & 0.259 \\ 
            OCR & ICDAR $\rightarrow$ mCLIP & 0.208 & 0.153 & 0.244 & 0.186 & 0.284 & 0.287 & 0.228 & 0.218 \\ 
            Speech & Whisper $\rightarrow$ mCLIP & 0.242 & 0.111 & 0.221 & 0.290 & 0.267 & 0.214 & 0.320 & \textbf{0.321} \\ 
            Text & Description $\rightarrow$ mCLIP & \textbf{0.343} & 0.173 & 0.260 & 0.227 & 0.335 & 0.443 & 0.302 & 0.211 \\ 
            \bottomrule
        \end{tabular}
        \caption{}
        \label{tab:event_type}
    \end{subtable}
    \caption{\textbf{Breakdown of single-modality pipeline results.} Table~\ref{tab:language} shows that Chinese and Korean videos are particularly challenging, especially for vision-only models. From Table~\ref{tab:video_type}, extracted text is most beneficial for professional news broadcasts, while real-time raw footage remains challenging for all direct video content. Table~\ref{tab:query_type} demonstrates that queries based on specific modalities are generally best handled by the corresponding modality. Finally, from Table~\ref{tab:event_type}, visual content is most effective for most MultiVENT 1.0 event types.}
    \label{tab:breakdown_results}
\end{table*}

\begin{itemize}
    \item \textbf{VALOR} combines three encoders for single-modality representations with a decoder designed for multimodal text generation~\cite{chen2023valorvisionaudiolanguageomniperceptionpretraining}.
    \item \textbf{VAST} utilizes omni-modality pretraining to simultaneously fine-tune text, audio, and image encoders to enhance cross-modality learning~\cite{chen2023vastvisionaudiosubtitletextomnimodalityfoundation}.
    \item \textbf{InternVid 2} employs two expert models for video token-level unmasking during training and currently is state-of-the-art for zero-shot retrieval on \textsc{MSR-VTT}~\cite{wang2024internvideo2scalingfoundationmodels}.\footnote{\url{https://paperswithcode.com/sota/zero-shot-video-retrieval-on-msr-vtt}}
    \item \textbf{Language-Bind} leverages a \textsc{CLIP}-based encoder for all non-text modalities and has achieved state-of-the-art results on several video-understanding tasks~\cite{zhu2024languagebindextendingvideolanguagepretraining}.
\end{itemize}

Beyond VLMs, we also evaluate several baseline pipeline approaches that utilize information from a single modality.  In all cases, these system leverages \textsc{mCLIP}'s multimodal embedding space to efficiently compare model outputs to queries.

\begin{itemize}
    \item \textbf{mCLIP}: We first extract ten keyframes by detecting significant scene changes within a video and then extracting the midpoint frame.\footnote{\url{https://www.scenedetect.com}} The frames are passed through \textsc{mCLIP}'s image encoder, and the pooled embeddings are then compared to analogous query embeddings.
    \item \textbf{ICDAR} OCR: Using the same frames from the vision baseline, we extract all visible embedded text using a state-of-the-art multilingual OCR system~\cite{etter2023hybrid}, and pass the text output through mCLIP's text encoder.
    \item \textbf{Whisper} OCR: We utilize \textsc{Whisper}, a robust automatic speech recognition system~\cite{radford2022whisper}, to transcribe each video's audio track, and the resulting text is passed through \textsc{mCLIP}'s text encoder.
    \item \textbf{Description}: This baseline embeds each video’s human-written description through \textsc{mCLIP}'s text encoder.
\end{itemize}

Table \ref{tab:baselines} presents the performance of these baselines on \textsc{MultiVENT 2.0}, as well as results on \textsc{MSR-VTT}, a standard benchmark video retrieval datasets. A key observation is that while VLMs excel on prior collections, most perform poorly on our task. This is likely due to two main factors: the length of our videos, as VLMs are typically trained on short video segments; and the significant domain mismatch between simple visual concepts and complex event-based natural language, the latter being much harder to map directly to visual features. This second issue is compounded by some \textsc{MultiVENT 2.0} queries targeting non-visual aspects of the video.

Another notable takeaway is that single-modality pipeline systems show promise, though there is considerable room for improvement. Table~\ref{tab:breakdown_results} highlights how different aspects of the problem can be better addressed by specific modalities. The distinction is particularly evident in the breakdown of video types in Table~\ref{tab:video_type}: speech is highly effective for retrieving professional news broadcasts, yet it, along with embedded text, prove to be almost useless with raw content. In contrast, human-written descriptions and visual content are more robust across video types, with descriptions being the most useful for retrieving raw footage.

Based on these findings, a key question moving forward is whether to allow the use of text metadata. This may be appropriate in certain settings, but given the rapid rise of online visual content, it is unrealistic to assume that clean, human-written descriptions of a video will always be available. This is especially true for raw, user-generated content, which is often posted in real-time as events unfold. As a result, we propose two versions of the final evaluation task for future systems: one where descriptions are disallowed (\textsc{MultiVENT Test-noDesc}) and one where they are permitted (\textsc{MultiVENT Test-Desc}).

\section{Conclusion}
\label{sec:conclusion}

In this paper, we introduce \textsc{MultiVENT 2.0}, a massive multilingual, event-centric video retrieval collection that significantly broadens the scope and complexity of multimodal retrieval tasks. With over 218,000 videos and 3,900 queries targeting a diverse range of world events across six languages, \textsc{MultiVENT 2.0} poses substantial new challenges for vision-language models. Our results indicate that while pre-trained models such as InternVid and VAST have achieved impressive performance on prior retrieval collections, they face considerable difficulties in this setting, largely due to the domain mismatch between interpreting simpler visual concepts and the complex multimodal nature of event-based content. Pipeline systems incorporating specialized single-modality models show promise for subsets of the task, but these remain insufficient in isolation. In addition, retrieving raw video footage without the accompanied human-written text descriptions is extremely challenging, a limitation especially relevant in real-time event scenarios. Our results highlight the need for more robust multimodal retrieval systems, as exciting visual understanding and generation tasks cannot be leveraged in practice without first being able to accurately and efficiently triage relevant visual content.
{
    \small
    \bibliographystyle{ieeenat_fullname}
    \bibliography{main}
}

\clearpage
\setcounter{page}{1}
\maketitlesupplementary


\section{Query Creation - Annotation Guidelines}
\label{sec:queryguidelines}

In this section, we provide the guidelines given to our professional linguists for the task of writing queries targeting aspects of new events found within our larger collection sampled from InternVid, described in Section \ref{sec:query}. This annotation task was set up and run in Label Studio, a flexible data labeling platform that allows for multimodal annotations.\footnote{\url{https://labelstud.io}} When writing queries for \textsc{MultiVENT-Train} and \textsc{MultiVENT-Test}, we ran two iterations of this task: the first looked at a random sample of 200 Arabic, Chinese, English, Korean, and Russian videos. For \textsc{MultiVENT-Test} we added Spanish as a sixth language. We focused on annotating InternVid videos from the News \& Politics and Sports domains, as these were the categories with the most event-focused content. 

After this initial round, there was not sufficient raw event-focused videos. To address this, we again sampled 200 videos per language from InternVid, this time filtering out any videos with standard resolutions, and continued until we had at least 20 annotated videos for each language/video type pair.

\subsection{Task Introduction}

The goal of this task is to annotate videos based on event templates to identify relevant information/entities across multiple modalities (text descriptions, video footage, text in videos, and audio/speech). These annotations were used to create queries based on the \textsc{MultiVENT 1.0} dataset and the events it contains. For the retrieval task to be meaningfully challenging, the set of documents from which systems are retrieving needs to be quite large (in the hundreds of thousands). As such, researchers have supplemented the \textsc{MultiVENT} dataset with videos from a larger collection of videos, \textsc{InternVid}, and have begun writing queries associated with videos from that set.

In this annotation task, language experts will write queries for \textsc{InternVid} videos, with a focus on queries that come from specific modalities. For each annotation, you will be presented with a video (from YouTube) and its description (if present). Your task will be to watch the video, read the description, and write up to 4 English search queries based on information in the video, video description, video audio, and video text.

\subsection{Video Relevance}

\paragraph{What is an event?} We broadly define an ``event" as some set of properties and their changes (or lack of change) over time. A single, distinct event can only exist at one set location(s) across one set time interval. For this task, we generally constrain an event to something that occurs naturally, that is, not a staged set of predetermined actions (think movies, tutorial videos, etc.). We generally want to write queries that pertain to the main events depicted in the videos. 

\paragraph{When should I skip a video?} If the video falls into one of the below categories, select \textit{Not Relevant} and submit the annotation. 

\begin{itemize}
    \item If it is hard to tell what the main event in the video is.
    \item If there is minimal event-relevant information present in the video (for example, a black screen, or a slideshow of irrelevant images). 
    \item If the video is staged or otherwise not natural (movies, commercials, tutorial videos, etc.). 
    \item If the video is not mostly in the language you are annotating for (this applies to the description as well as the audio content). 
    \item If you are uncomfortable watching the video. Please do not annotate any videos that you don’t want to watch!
\end{itemize}

\subsection{Video Classification and Video Type}

Select an event type that best describes the event in your video. If none of the listed event types apply, select \textit{Other event type}.

\begin{itemize}
    \item Disaster (fire, earthquake, hurricane, etc.)
    \item Political Election
    \item Political Protest
    \item Other Political Development (sanctions, treaty, leader death, etc.)
    \item Social Event (festival, convention, celebration, etc.)
    \item Sporting Event
    \item Technical/Scientific Launch or Discovery
    \item Other Event Type
\end{itemize}

\text{}\\Then, indicate the type of video:
\begin{itemize}
    \item Professional news broadcast
    \item Non-professional edited footage
    \item Mostly raw footage
    \item Raw footage
\end{itemize}

\text{}\\The guidelines for video type classification are: 
\begin{itemize}
    \item The video is \textbf{professional footage} if you can imagine the clip appearing on broadcast TV. For example, there is an anchor/reporter, there are news logos and text banners with headlines, etc. 
    \item The video is \textbf{non-professional edited footage} if it is not professional AND there is superimposed OCR/graphics, other visual special effects have been applied, or there are multiple clips spliced together. For example, many YouTube streamers’ videos are “edited non-professional.”
    \item The video is \textbf{mostly raw footage} if the video is a mostly continuous raw stream of content but has minor additional content (i.e., one of speech/music overlay, text overlay, or minimal scene cuts). 
    \item The video is \textbf{true raw footage} if it is a continuous, unedited stream of footage.  
\end{itemize}

\subsection{Query Writing}

Tips for good query writing:
\begin{itemize}
    \item All of the queries should be in English. 
    \item Write queries in the style you would use for a search engine, e.g., Google. The queries should be short and do not need to be written in complete sentences. 
    \item Only write queries based on information that would help you learn more about the event in question. There may be video, audio, or text content that is not relevant to the main event topic—these irrelevant details should not be the basis for your queries. 
    \item The queries should be unique from one another. If you cannot write a query for unique information in a particular modality, check the relevant box for why you left the query blank (see below). 
    \item You can use supplemental internet searches to clarify information contained in the video description, e.g., to check spelling for a person’s name. 
    \item There are multiple ``right answers" for each video’s queries.  
\end{itemize}

\paragraph{}Each query has an accompanying text box; write your query and hit Enter to submit that query. Once submitted, the query will appear in a green text box. 

\subsubsection{Base Event Query}

\begin{figure}[ht!]
    \centering
    \centering
    \includegraphics[width=0.47\textwidth]{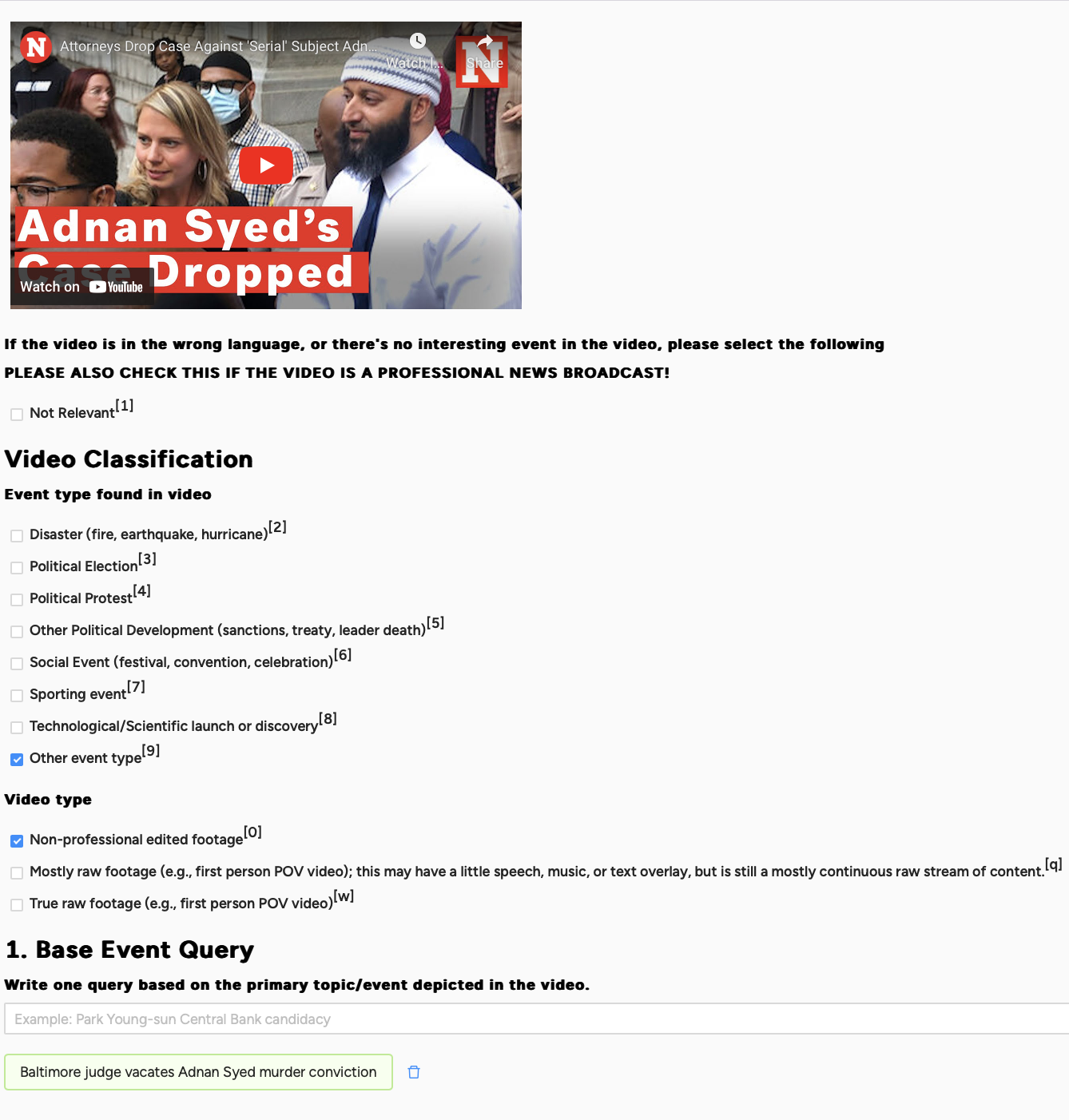}
    \caption{Example and Label Studio interface for video classification annotations and base query creation i.e., the phrase that represents the ``Wikipedia title" of a current event.}
    \label{fig:base}
\end{figure}

Watch the video. You may also read the title/description to get a better sense of what the video is depicting. Then, write one query based on the \textbf{primary topic/event} depicted in the video. This should include general information about what happened. The goal is to capture specific events. Each query should be approximately 3-7 words. An example of this question is provided in Figure~\ref{fig:base}.

\subsubsection{Text-based Query}
\begin{figure*}[htbp]
    \centering
    \begin{subfigure}[b]{\textwidth}
        \centering
        \includegraphics[width=0.97\textwidth]{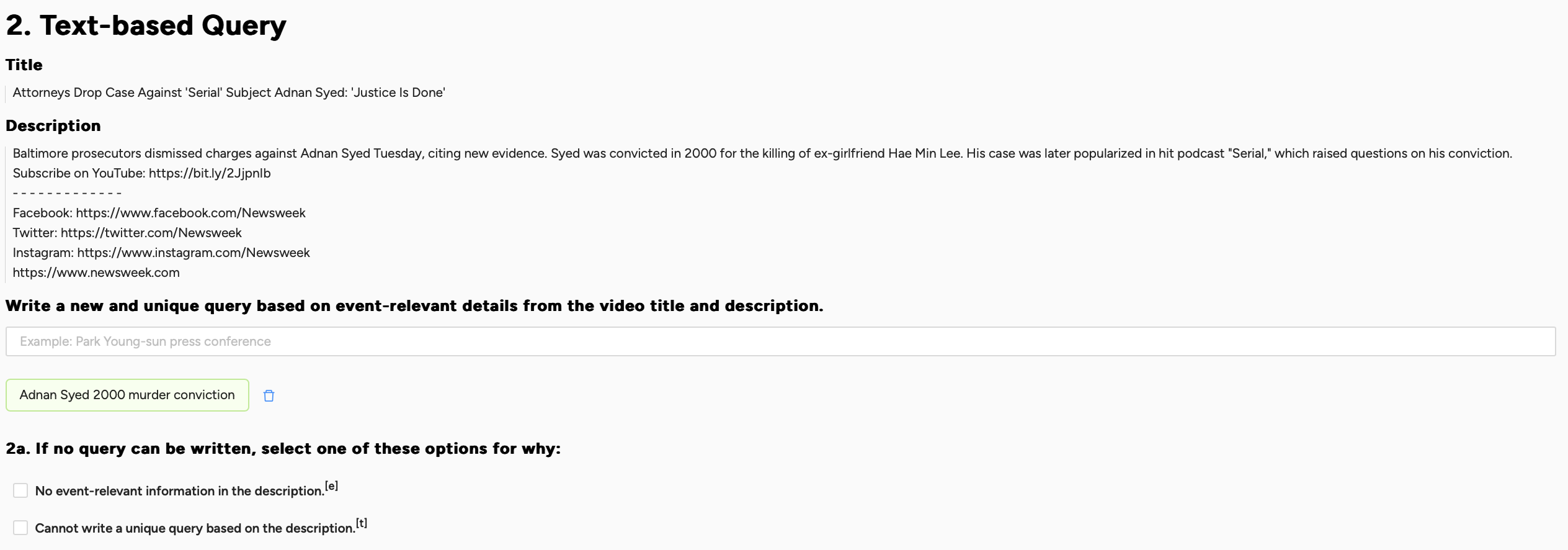}
        \caption{Query creation using only the associated text description}
        \label{fig:text}
    \end{subfigure}
    \vspace{0.1cm}
    \begin{subfigure}[b]{\textwidth}
        \centering
        \includegraphics[width=0.97\textwidth]{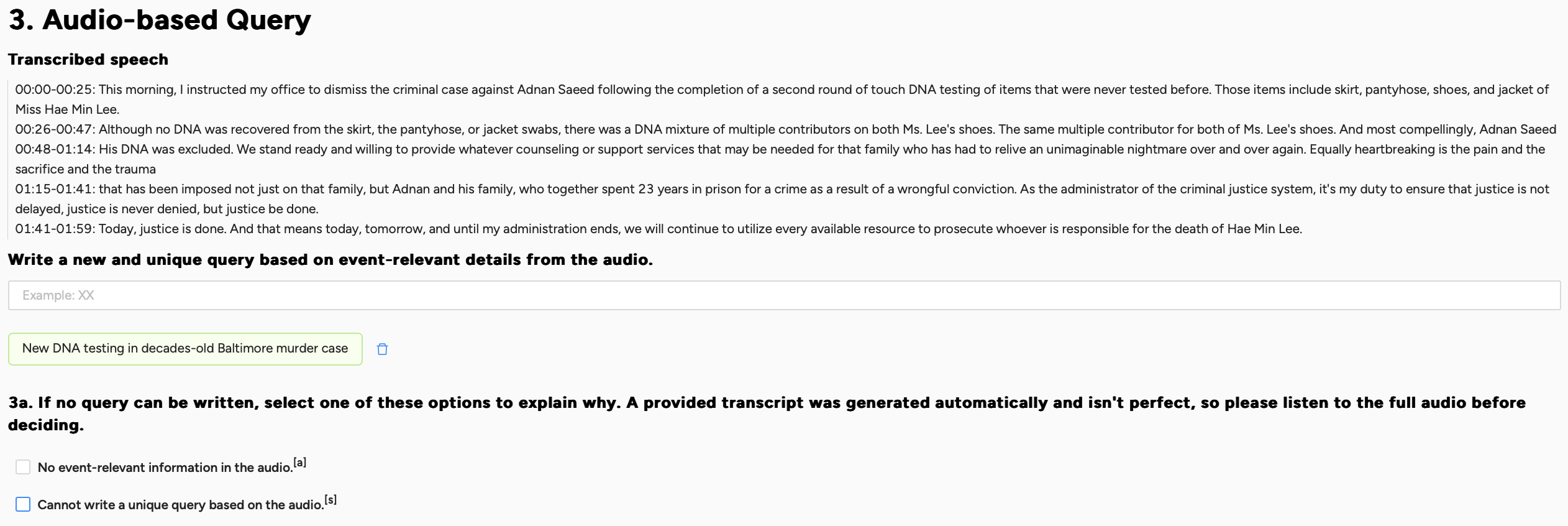}
        \caption{Query creation process using only audio, with associated transcribed text for assistance}
        \label{fig:audio}
    \end{subfigure}
    \vspace{0.1cm}
    \begin{subfigure}[b]{\textwidth}
        \centering
        \includegraphics[width=0.97\textwidth]{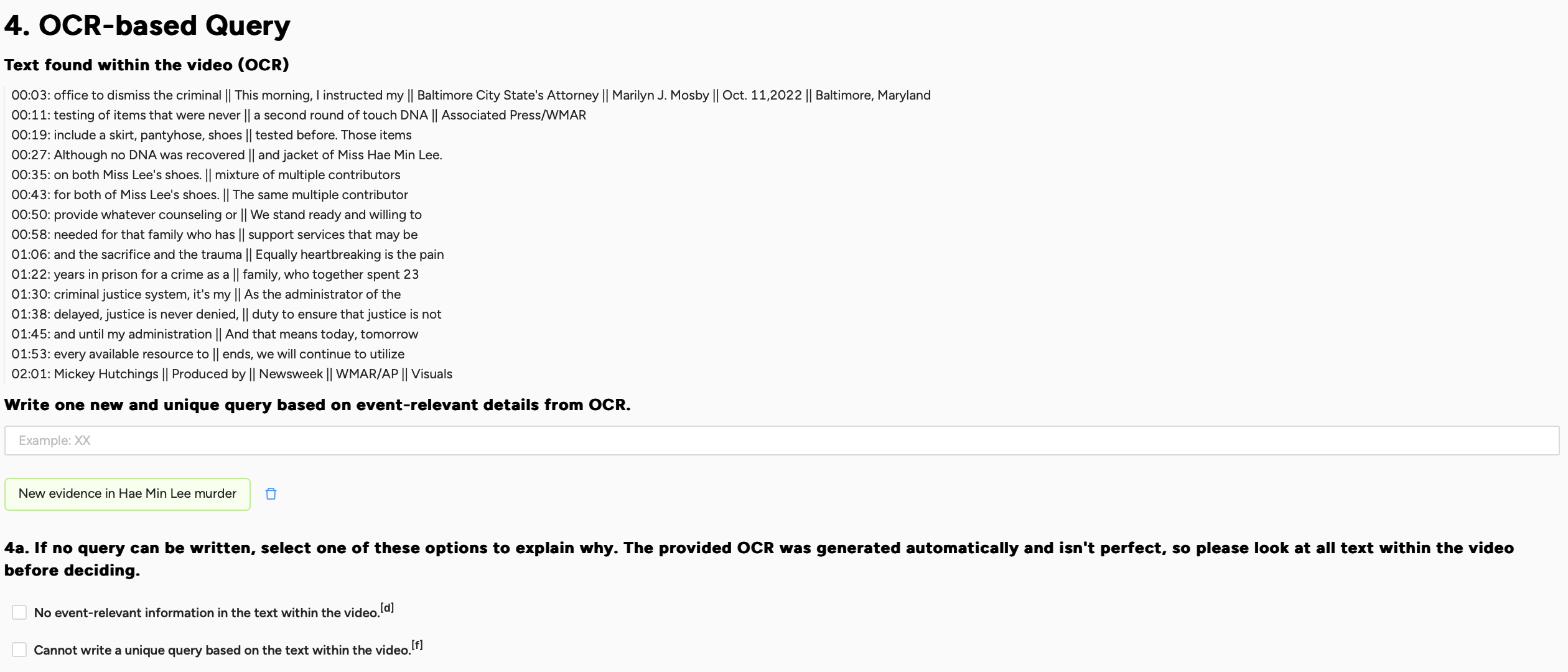}
        \caption{Query creation process using only embedded text, with associated OCR output for assistance}
        \label{fig:ocr}
    \end{subfigure}
\label{fig:queryannotation}
\caption{Example and Label Studio interface for writing queries targeting specific aspects of events. For each question, we ask annotators to only use the text description, audio, and embedded text, respectively.}
\end{figure*}

Write one query based on event-relevant details from the video description. This query should cover information that is not already addressed by previous queries. The query should be approximately 3-7 words. An example of this question and its associated interface is provided in Figure~\ref{fig:text}.

If no unique query can be written, select an option explaining why:
\begin{itemize}
    \item No event-relevant information in the description.
    \item Cannot write a unique query based on the description.
\end{itemize}

\subsubsection{Audio-based Query}

Write one query based on event-relevant details from the video audio/speech. This query should cover information that is not already addressed by previous queries. The query should be approximately 3-7 words. 

To aid annotation, transcribed speech has been provided (in the original language or translated, depending on the project you are working on). Note that these transcriptions, especially once translated, can have errors. Listen to the audio if you are a language expert and try to confirm spellings of names, places, etc. before finalizing your queries. An example of this question and its associated interface is provided in Figure~\ref{fig:audio}.

If no unique query can be written, select an option explaining why:
\begin{itemize}
    \item No event-relevant information in the audio.
    \item Cannot write a unique query based on the audio.
\end{itemize}

\subsubsection{OCR-based query}

Write one query based on event-relevant details from text in the video footage (e.g., signs, captions, etc.). Read the text found within the video (\textit{OCR}) section. This query should cover information that is not already addressed by previous queries. The query should be approximately 3-7 words.

To aid annotation, OCR transcripts have been provided (in the original language or translated, depending on the project you are working on). Note that these transcriptions, especially once translated, can have errors and omissions. Consult the video footage for text that may have been missed or misrendered by OCR and consider taking a screenshot and re-translating in Google Translate if you are not a language expert before finalizing your queries. An example of this question and its associated interface is provided in Figure~\ref{fig:ocr}.

If no unique query can be written, select an option explaining why:
\begin{itemize}
    \item No event-relevant information in the text within the video.
    \item Cannot write a unique query based on the text within the video.
\end{itemize}

\section{Relevance Judgment - Annotation Guidelines}
\label{sec:relevanceguidelines}

In this section, we provide the guidelines given to our professional linguists to judge the relevance of previously unseen query/video pairs, described in Section \ref{sec:relevance}.

\subsection{Introduction}

The goal of the task is for models to differentiate, i.e., through a ranked list, between relevant and irrelevant documents with respect to each query. Using Google as an example use case, a user expects the content most relevant to their search to appear in the first page of results. Through our prior query creation, we have developed a set of over 3,900 unique queries. To make the retrieval task meaningfully difficult, we have also included a large set of distractor videos (videos for which we have not written queries) in the overall document set. 

When evaluating models on the retrieval task, our metrics assume that any given unseen video will be \textit{Not Relevant} for any given query. However, it is possible that there are videos in our distractor set that are relevant for some of our queries. We want to find as many relevant videos as possible for a query to reduce the possibility of accidentally labeling something not relevant that actually is (called a \textit{false negative}). This means potentially increasing the number of videos that are judged as relevant. This will help us more accurately judge systems. 

Providing Relevance Judgment annotations involves assessing a query-video pair and determining if, for that query, the video is \textit{Very Relevant}, \textit{Somewhat Relevant}, or \textit{Not Relevant}. The sections below provide more detailed guidance for this task. 

\subsection{Annotation Instructions}

For each annotation task, you will be presented with a video (from YouTube), its title and description (if present), and extracted speech and embedded text. The text will be translated to English, if necessary (if you are a language expert and want to see the original title/description, open the video in a separate window in YouTube. 

\subsubsection{Video Type}

After watching the video, indicate the type of video—professional footage/news broadcast, non-professional edited footage, or raw footage (i.e., first person POV video). An example of this component is shown in Figure~\ref{fig:relevance_example}. \\\\
The guidelines for video type classification are: 
\begin{itemize}
    \item The video is \textbf{professional footage} if you can imagine the clip appearing on broadcast TV. For example, there is an anchor/reporter, there are news logos and text banners with headlines, etc. 
    \item The video is \textbf{non-professional edited footage} if it is not professional AND there is superimposed OCR/graphics, other visual special effects have been applied, or there are multiple clips spliced together. For example, many YouTube streamers’ videos are “edited non-professional.”
    \item The video is \textbf{mostly raw footage} if the video is a mostly continuous raw stream of content but has minor additional content (i.e., one of speech/music overlay, text overlay, or minimal scene cuts). 
    \item The video is \textbf{true raw footage} if it is a continuous, unedited stream of footage. 
\end{itemize}

\paragraph{}Note: some news agencies/platforms release a mix of video types. A “non-professional edited” video may be from a professional new company, but not meet our criteria for professional footage. 

\subsubsection{Relevance Judgments - Overview}

Next, you will see a list of 1 or more English queries that you will judge that video against for relevance. For each query, you will select one of the following options: 
\begin{itemize}
    \item \textbf{Not Relevant}
    \item \textbf{Possibly Relevant}, i.e., the query mentions a clear event, and the video is possibly about that event 
    \item \textbf{Partially Relevant}, i.e., the query mentions a clear event, the video is about that event, but does not show everything in the query 
    \item \textbf{Very Relevant}
\end{itemize}

\paragraph{}You will also have a text box to optionally leave comments (e.g., if you are not sure and/or escalating the judgment and want to briefly explain your reasoning). 

\subsubsection{About Our Queries}

\paragraph{Base event queries} When we wrote the base event queries, we tried to identify the main event being addressed by a given video. The model for this approach was MultiVENT base event queries, which were based on trending events that i.e. were covered by a Wikipedia article. This means that often the base event queries are broader than the videos they were written for, which may only cover specific aspects of the event in question. Some examples are below:
\begin{itemize}
    \item A video covering loosening mask requirements in March 2022 in Canada might have the base event query \textit{Covid-19 pandemic in Canada}.
    \item A video depicting a post-game interview with Chiefs quarterback Patrick Mahomes might have the base event query \textit{Super Bowl LVIII}.
    \item A video of highlights from a presidential debate might have the base event query \textit{2022 South Korean presidential election}.
\end{itemize}

\paragraph{}For each of the query-video pairs above, the video would be Very Relevant for that query, even though it may not address the event as a whole. The base event queries may also name the event using details not explicitly mentioned in the video (for example, the year). However, if you are able to determine that the event described in the query is the same underlying event that the video is about, that video will be \textit{Very Relevant}. 

\begin{figure*}[htbp]
    \centering
    \begin{subfigure}[b]{\textwidth}
        \centering
        \includegraphics[width=0.79\textwidth]{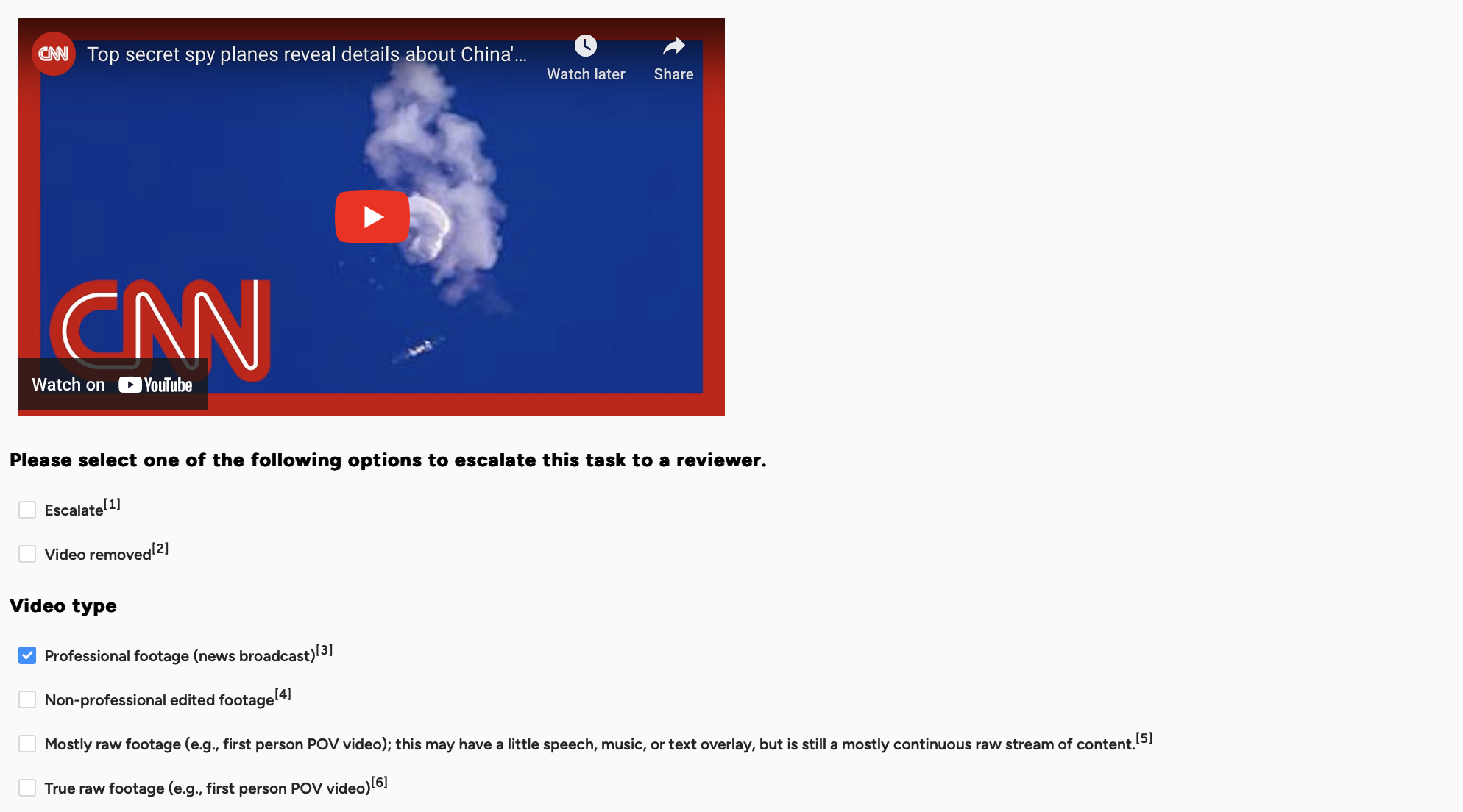}
        \caption{Example video and classification annotations}
        \label{fig:relevance_example}
    \end{subfigure}
    \vspace{0.1cm}
    \begin{subfigure}[b]{\textwidth}
        \centering
        \includegraphics[width=0.79\textwidth]{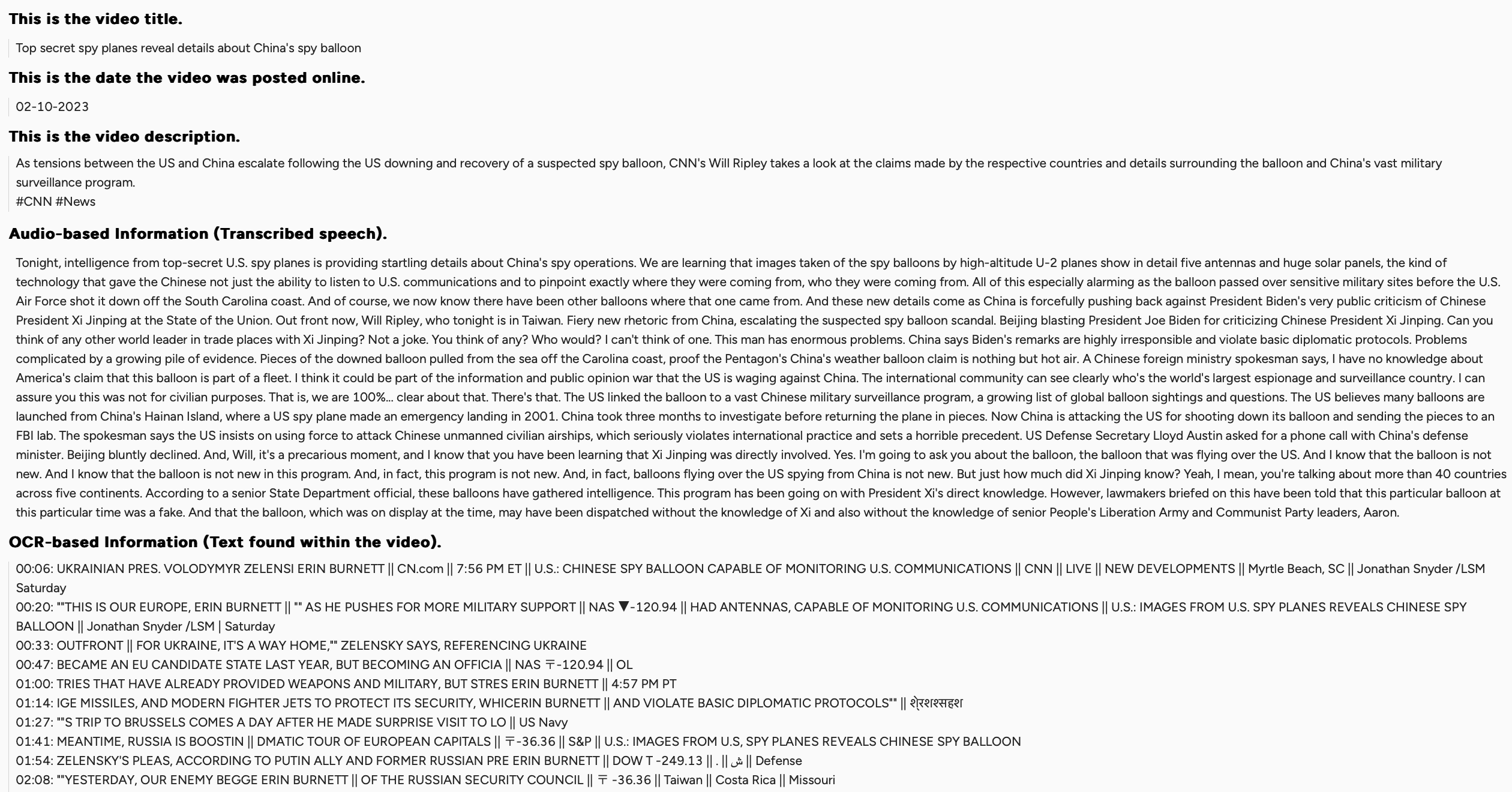}
        \caption{Provided text extracted from the video}
        \label{fig:relevance_text}
    \end{subfigure}
    \vspace{0.1cm}
    \begin{subfigure}[b]{\textwidth}
        \centering
        \includegraphics[width=0.79\textwidth]{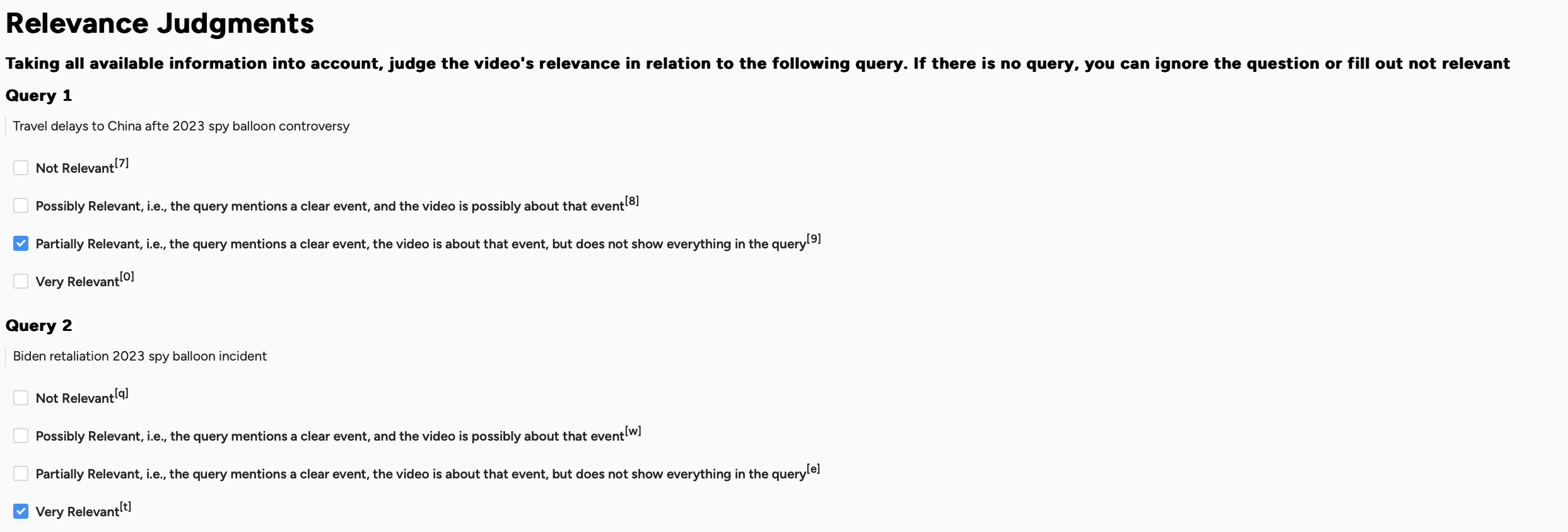}
        \caption{Queries for which the video is judged for relevance.}
        \label{fig:relevance_judgments}
    \end{subfigure}
\label{fig:relevance_annotation}
\caption{Example of the annotation process for judging relevance for previously unseen query/video pairs.}
\end{figure*}

\paragraph{Specific Queries} We wrote specific queries based on details found in videos. Specific queries will usually refer to the event in question as well as a specific aspect of that event. When you judge relevance for specific queries, you will want to consider (to the extent that they are decomposable) both the event described by the query and the specific aspect(s) or detail(s) about the event included in the query. Some examples are below:
\begin{itemize}
    \item For a query \textit{Black Sea drone incident damaged propeller}, you will want to consider 1) is the video about the Black Sea drone incident and, if yes, 2) does the video also include reference (in the image, description, speech, and/or embedded text) to a damaged propeller? 
    \item For a query \textit{Democratic Party response to Daejang-dong scandal}, you will want to consider 1) is the video about the Daejang-dong scandal, and 2) does the video also include content about the Democratic Party’s response to the scandal? 
\end{itemize}

\paragraph{}If a specific query does not explicitly reference the \textit{base event}, evaluate relevance based on each detail in the query. Some examples are below:
\begin{itemize}
    \item For a query \textit{Eric Adams hiring Director of Rat Control}, consider whether the video explicitly refers to Eric Adams (or New York Mayor) and to Director of Rat Control/Director of Rodent Mitigation. 
    \begin{itemize}
        \item If the video is about Eric Adams but unrelated to rat control, the video is Not Relevant. 
        \item If the video is about hiring the Director of Rat Control but does not refer to Eric Adams/the New York Mayor, the video is Partially Relevant. 
    \end{itemize}
    \item For a query \textit{Canada eases entry restrictions September 2022}, consider whether the video is about updates to entry restrictions, in Canada, in September 2022. 
    \begin{itemize}
        \item Because the query points to a specific event, if the video is about Canadian entry restrictions not in September 2022, the video is Not Relevant. 
        \item If the video is about Canadian entry restrictions and the date/time period is not clear, the video is Partially Relevant. 
    \end{itemize}
\end{itemize}

\subsubsection{Consulting Outside Sources}

If you are unfamiliar with the event, people, places, etc. described by a given query, we encourage you to search for the event and/or entities online to learn more about the query’s context. For named entities, this is particularly helpful, because the queries might refer to those entities in different ways than the videos do. For example, a query might say \textit{Norilsk avalanche} and a video might refer to an \textit{avalanche in the Krasnoyarsk region}; if you are not familiar with the region, an internet search may be required to determine that the video is relevant for that query. 

When assessing the queries and videos for relevance, try to think conceptually about what a query is asking for, rather than looking for that exact wording in the video. 

\subsubsection{About Our Videos and Metadata}

Most videos you will see for this task are from YouTube. You may also encounter some from Twitter. In addition to the video, you will also see various metadata for that video, shown in Figure \ref{fig:relevance_text}. For videos not originally in English, the metadata will have been translated into English.

\begin{itemize}
    \item \textbf{Video Title + Video Description.} If you are a language expert and want to see the title and description in the original language, open the video in another window. 
    \item \textbf{Posted Date.} Note that while it is helpful to use this date as a clue for the event depicted by the video, the date the video was posted online will not necessarily correspond closely to the date of the event. Consider the date along will all other information available to you.
    \item \textbf{ASR (automated speech recognition) output.} Speech from the video has been automatically transcribed, then translated into English if necessary. 
    \item \textbf{OCR (optical character recognition) output.} A sample of text from the video (i.e., captions, banners, signs) has been automatically transcribed, then translated into English if necessary. Not all video frames were considered for this OCR output—this means that there may be text that was not caught at all for transcription and translation. You can supplement this information by pausing the video on a frame that has text on it, taking a screenshot of the video frame, and pasting the image into Google Translate for images.    
\end{itemize}

Each automated process described above has the potential for errors. This is especially true for the recognition and translation of proper nouns, e.g., names.  Consider all of the information available in the video, metadata, and outside sources to judge whether details in the video correspond to details from the query. 

\subsubsection{Relevance Judgments - Detailed Guidance}

A video is Relevant with respect to query if some content in the video addresses the information-seeking need defined by the query. How primary that relevant content is within the video (e.g.,, if it is the main topic of the videos vs. If it is mentioned in passing) does not affect the video’s relevance for the query.

Each query defines what kind of information we are looking for in the video (see discussion of Base Event queries and Specific queries above). Some videos will be Very Relevant (i.e., they completely address all aspects of a query), others will be Not Relevant (they do not address what the query is looking for). There are also two intermediate relevance judgments, Partially Relevant and Possibly Relevant. Example queries are provided in Figure \ref{fig:relevance_judgments}, and more details about each category are below:

\paragraph{Very Relevant [3]} A video is \textit{Very Relevant} for a query if it addresses all components of a query. 
\begin{itemize}
    \item For a base event query, a \textit{Very Relevant} video is one about that event. 
    \item For a specific query, a \textit{Very Relevant} video is one that is about the event described by the query (if applicable) and includes the specific information/details in the query. 
\end{itemize}

\paragraph{Partially Relevant [2]} A video is \textit{Partially Relevant} for a query if it addresses the general topic of the query but does not include the specific detail(s) the query is asking for. 
\begin{itemize}
    \item For a specific event query, a \textit{Partially Relevant} video is one that covers the main event/topic in the query but is missing a specific detail included in the query.
    \item If you can confirm that a query is about an event distinct from the one in the query, the video is Not Relevant.
\end{itemize}

\paragraph{Possibly Relevant [1]} A video is \textit{Possibly Relevant} if it is unclear that the video is relevant for the query. 
\begin{itemize}
    \item For a base or specific event query, a \textit{Possibly Relevant} video is one that may cover an aspect of the event, but it is not entirely clear that it does so (i.e., it is possible that the video is about the event/topic, but it could also be about a different event.). 
    \item If you can confirm that a query is about an event distinct from the one in the query, the video is \textit{Not Relevant}. 
\end{itemize}

\paragraph{Not Relevant [0]} A video is \textit{Not Relevant} if it does not address the event described in the query. This includes videos that are about similar but distinct events—for example, a video about the \textit{2018 Winter Olympics} when the query is about the \textit{2022 Winter Olympics}. 

\paragraph{When to escalate} Escalate in the following scenarios: 

\begin{itemize}
    \item You are unable to access or play a video (before you escalate for this reason, try to open the video in another window). [Select \textit{Video Removed}] 
    \item The video is in the wrong language (in relation to the project you are working on) 
    \item Given your understanding of the task and these guidelines, you are unable to determine the video’s relevance in relation to one or more queries. In this scenario, please also post a message to the Teams chat describing your question. If your question is answered, you can return to the task later, update judgments as needed, and un-select the Escalate option. [Select \textit{Escalate}] 
\end{itemize}

\paragraph{}Note that you can still provide relevance judgments for some queries, even if you escalate the task as a whole. If there are queries for which you can confidently judge relevance, please do so. If you escalate a task, leave a brief explanation in the comment box to describe your reason for escalation and any other details you think would be helpful for the next person evaluating the video.

\end{document}